\documentclass{article}

\usepackage{arxiv}

\usepackage[utf8]{inputenc} % allow utf-8 input
\usepackage[T1]{fontenc}    % use 8-bit T1 fonts
\usepackage{hyperref}       % hyperlinks
\usepackage{url}            % simple URL typesetting
\usepackage{booktabs}       % professional-quality tables
\usepackage{amsfonts}       % blackboard math symbols
\usepackage{nicefrac}       % compact symbols for 1/2, etc.
\usepackage{microtype}      % microtypography
\usepackage{xcolor}         % colors

\usepackage[american]{babel}
\usepackage{amsmath}
\usepackage{amssymb}
\usepackage{amsthm}
\usepackage{tikz}
\usetikzlibrary{arrows,cd,decorations.markings,shapes.geometric,shapes}
\usepackage{mathtools}
\usepackage{ebproof}

\usetikzlibrary{arrows,cd,decorations.markings,shapes.geometric,shapes}

\usepackage{tikzit}
\usepackage[all,cmtip]{xy}

\usepackage{tikzit}

\usetikzlibrary{circuits.ee.IEC}

\usetikzlibrary{calc,arrows.meta}

\newcommand\cofib\rightarrowtail

\newcommand\mdel[1]{}

\usepackage{txfonts}
\usepackage{pxfonts}

\makeatletter
\newcommand{\xdashrightarrow}[2][]{\ext@arrow 0359\rightarrowfill@@{#1}{#2}}
\newcommand*{\doublerightarrow}[2]{\mathrel{
  \settowidth{\@tempdima}{$\scriptstyle#1$}
  \settowidth{\@tempdimb}{$\scriptstyle#2$}
  \ifdim\@tempdimb>\@tempdima \@tempdima=\@tempdimb\fi
  \mathop{\vcenter{
    \offinterlineskip\ialign{\hbox to\dimexpr\@tempdima+1em{##}\cr
    \rightarrowfill\cr\noalign{\kern.5ex}
    \rightarrowfill\cr}}}\limits^{\!#1}_{\!#2}}}
\newcommand*{\triplerightarrow}[1]{\mathrel{
  \settowidth{\@tempdima}{$\scriptstyle#1$}
  \mathop{\vcenter{
    \offinterlineskip\ialign{\hbox to\dimexpr\@tempdima+1em{##}\cr
    \rightarrowfill\cr\noalign{\kern.5ex}
    \rightarrowfill\cr\noalign{\kern.5ex}
    \rightarrowfill\cr}}}\limits^{\!#1}}}
\makeatother

\makeatletter
\newcommand{\twoarrows}[3][0.2ex]{%
  \mathrel{\mathpalette\twoarrows@{{#1}{#2}{#3}}}%
}
\newcommand{\twoarrows@}[2]{\twoarrows@@#1#2}
\newcommand{\twoarrows@@}[4]{%
  \vcenter{\offinterlineskip\m@th
    \ialign{\hfil##\hfil\cr
      $#1#3$\cr
      \noalign{\vskip#2}
      $#1#4$\cr
    }%
  }%
}
\makeatother

\newcommand{\beq}{\begin{equation}}
\newcommand{\eeq}{\end{equation}}

\newtheorem{theorem}{Theorem}
\newtheorem{definition} {Definition} 
\newtheorem{lemma}{Lemma}

\usepackage[boxruled]{algorithm2e}
\usepackage{algorithmic}

\usepackage{multirow}
\usepackage{booktabs}
\usepackage{balance}
\usepackage{subfig}

\DeclareFontFamily{U}{dmjhira}{}
\DeclareFontShape{U}{dmjhira}{m}{n}{ <-> dmjhira }{}

\usepackage{natbib} % has a nice set of citation styles and commands
\usepackage{mathtools} % amsmath with fixes and additions
\usepackage{booktabs} % commands to create good-looking tables
\usepackage{tikz} % nice language for creating drawings and diagrams

\title{Universal Reinforcement Learning in Coalgebras: Asynchronous Stochastic Approximation via Coinduction \thanks{Draft under revision.} }

\author{ Sridhar Mahadevan \\
	Adobe Research and University of Massachusetts, Amherst\\
	\texttt{smahadev@adobe.com, mahadeva@umass.edu}
}

\hypersetup{
pdftitle={A template for the arxiv style},
pdfsubject={q-bio.NC, q-bio.QM},
pdfauthor={David S.~Hippocampus, Elias D.~Striatum},
pdfkeywords={First keyword, Second keyword, More},
}

\begin{document}
\maketitle

\begin{abstract}
In this paper, we introduce a categorial generalization of RL, termed universal reinforcement learning (URL), building on powerful mathematical abstractions from the study of coinduction on non-well-founded sets and universal coalgebras, topos theory, and categorial models of asynchronous parallel distributed computation.  In the first half of the paper, we  review the basic RL framework, illustrate the use of categories and functors in RL, showing how they lead to interesting insights. In particular, we also introduce a standard model of asynchronous distributed minimization proposed by Bertsekas and Tsitsiklis, and describe the relationship between metric coinduction and their proof of the Asynchronous Convergence Theorem. The space of algorithms for MDPs or PSRs can be modeled as a functor category, where the co-domain category forms a topos, which admits all (co)limits, possesses a subobject classifier, and has exponential objects. In the second half of the paper, we move on to universal coalgebras.   Dynamical system models, such as Markov decision processes (MDPs),  partially observed MDPs (POMDPs), a predictive state representation (PSRs), and linear dynamical systems (LDSs) are all special types of coalgebras.  We describe a broad family of universal coalgebras, extending the dynamic system models studied previously in RL.  The core problem in finding fixed points in RL  to determine the exact or approximate (action) value function is generalized in URL to determining the final coalgebra asynchronously in a parallel distributed manner. 
\end{abstract}

% keywords can be removed

\newpage 

\tableofcontents

\newpage 

\section{Introduction}\label{sec:intro}

The computational study of reinforcement learning (RL) dates back to Arthur Samuel's pioneering work on the checkers player \citep{samuel1959}. Excellent textbook accounts of RL are given in \citep{bertsekas:rlbook,DBLP:books/lib/SuttonB98}.  RL has recently begun to play a key role in the fine-tuning of large language models (LLMs) \citep{deepseekai2025deepseekr1incentivizingreasoningcapability}. RL has also been successfully combined with deep learning \citep{deeplearningreview-2009}, leading to many interesting demonstrations on challenging video games. 

In this paper, we introduce a categorical generalization of RL, which we term universal reinforcement learning (URL), building on the powerful mathematical abstractions of category theory \citep{maclane:71,riehl2017category}, topos theory \citep{Johnstone:592033,maclane:sheaves}, universal coalgebras \citep{jacobs:book,rutten2000universal}, and formal models of parallel asynchronous distributed computation \citep{bertsekas:pdc,witsenhausen:1975}. The main thesis underlying URL can be stated as follows: 

\begin{quote}
{\em URL, and by specialization RL,  is essentially a process of stochastic metric coinduction in a universal coalgebra defined over a topos of (action)value functions, and all TD-type algorithms can be analyzed in terms of iterative procedures for discovering final coalgebras.}.
\end{quote}

It is the goal of this paper to explain justify this assertion more formally. Doing so will require significant background in modern category theory, and we can only provide the briefest of introductions here. Besides the wealth of available textbooks on category theory \citep{maclane:71,riehl2017category,richter2020categories}, topos theory \citep{maclane:sheaves,bell,goldblatt:topos,Johnstone:592033}, and universal coalgebras \citep{jacobs:book}, we refer the reader to our past paper on universal decision models \citep{sm:udm} for an introduction to categorical representations in RL. Our recent papers on categorial causality \citep{DBLP:journals/entropy/Mahadevan23,cktheory,mahadevan2025toposcausalmodels}, categorial foundations of generative AI \citep{mahadevan2024gaiacategoricalfoundationsgenerative}, universal imitation games \citep{mahadevan2024universalimitationgames}, and categorial large language models \citep{mahadevan2025rosesmellsweetcategorical} contain further examples on many applications of categorial concepts to AI and ML, to which we refer the reader. We will illustrate the main concepts in the context of RL in this paper, and emphasize intuition over rigor given the introductory tutorial nature of this paper. 

In essence, URL makes the following distinct contributions: 

\begin{enumerate}
    \item URL introduces the study of RL over universal coalgebras \citep{rutten2000universal}, which is interesting in that it brings with it a new notion of {\em state} \citep{jacobs:book} that is different from past work in RL. 

    \item URL broadens the family of dynamical system models studied previously in RL, to include a whole grammar of possible models defined as probabilistic or stochastic coalgebras \citep{SOKOLOVA20115095}. 

    \item URL introduces the concept of {\em diagrams} defined as functors $F: {\cal J} \rightarrow {\cal C}_{URL}$ as a way to specify properties of a URL category, such as (co)limits, pullbacks, pushouts, (co)equalizers etc., all of which are shown to play important roles in obtaining solutions. 

    \item URL introduces a new way to formulate stochastic approximation, which draw from categorical abstractions of previous work on asynchronous parallel distributed computation \citep{bertsekas:pdc,witsenhausen:1975}. 
\end{enumerate}

Our primary interest in this paper is bringing in new theoretical approaches to the study of RL, with the long-term goal of gaining insight into how to learn to make optimal decisions in a tractable way by exploiting structure. Most applications of large-scale RL require an enormous number of samples, and the scale of computation needed seems completely beyond what humans are capable of. We live only for a billion seconds or more, and yet many experimental studies in RL require tens of billions or more episodes. This discrepancy between what humans actually require to learn complex tasks, such as driving, from studies in RL show that the predominant challenge in RL remains how to specify a priori structure (or how to learn parts of such structure from previous tasks). 

Our work is also related to past work on MDP homomorphisms \citep{DBLP:conf/ijcai/RavindranB03} and PSR homomorphisms \citep{DBLP:conf/aaai/SoniS07}. These are special cases of the categorial structures studied in this paper, and will be used to illustrate our more abstract analysis. We hasten to add that our paper is not the first to study coalgebraic structure of MDPs. \citet{feys} describes the coalgebraic structure implicit in the definition of dynamic programming methods to solve MDPs, but does not study sampling-based RL methods in particular. \citet{kozen} explores the application of coalgebraic inference methods, such as {\em metric coinduction} in the Bellman equation as well. Both of these are specifically limited to the MDP model as well, whereas we explore a far broader range of coalgebras. 

Our work also relates to Witsenhausen's {\em intrinsic model} \citep{witsenhausen:1975}, which we explored previously in our work on universal decision models \citep{sm:udm}. Witsenhausen proposed the notion of an {\em information field}, a measure-theoretic notion of state, and also defined a taxonomy of decision problems based on constructing a topology of decision makers. In his framework, every decision-making instant was modeled as a separate decision maker, and his taxonomy included synchronous and asynchronous forms of decision making. There is of course an enormous literature in operations research on dynamic programming models, and we refer the reader to standard textbook treatments \citep{DBLP:books/lib/Bertsekas05}. Finally, our URL framework builds on past theoretical work on using stochastic approximation methods to analyze the convergence of RL algorithms, such as Q-learning \citep{kushner2003stochastic}. The main difference here is that we are using categorial constructions, such as Kan extensions \cite{kan}, to model martingale processes, building on recent work by \citet{rubenvanbelle:phd}. 

\section{Roadmap to the Paper} 

Here is a brief roadmap to our paper, which is comprised of two parts. 

\subsection{Part I: Functor Categories for RL}

A {\em category} ${\cal C}$  is a collection of {\em objects} $c \in {\cal C}$, and a collection of arrows between each pair of objects $f \in {\cal C}(c, c')$, including an identity arrow $1_c$ for each object $c \in {\cal C}$, and composability of arrows ${\cal C}(c,c') \times {\cal C}(c', c'') \rightarrow {\cal C}(c, c'')$.  A functor $F: {\cal C} \rightarrow {\cal D}$ maps each object $c \in {\cal C}$ to the object $Fc \in {\cal D}$ and each arrow $f \in {\cal C}$ to the arrow $Fc \in {\cal D}$. A category of functors ${\cal D}^{\cal C}$ is one where each object is a functor $F: {\cal C} \rightarrow {\cal D}$.  

\begin{figure}
    \centering
    \includegraphics[width=0.75\linewidth]{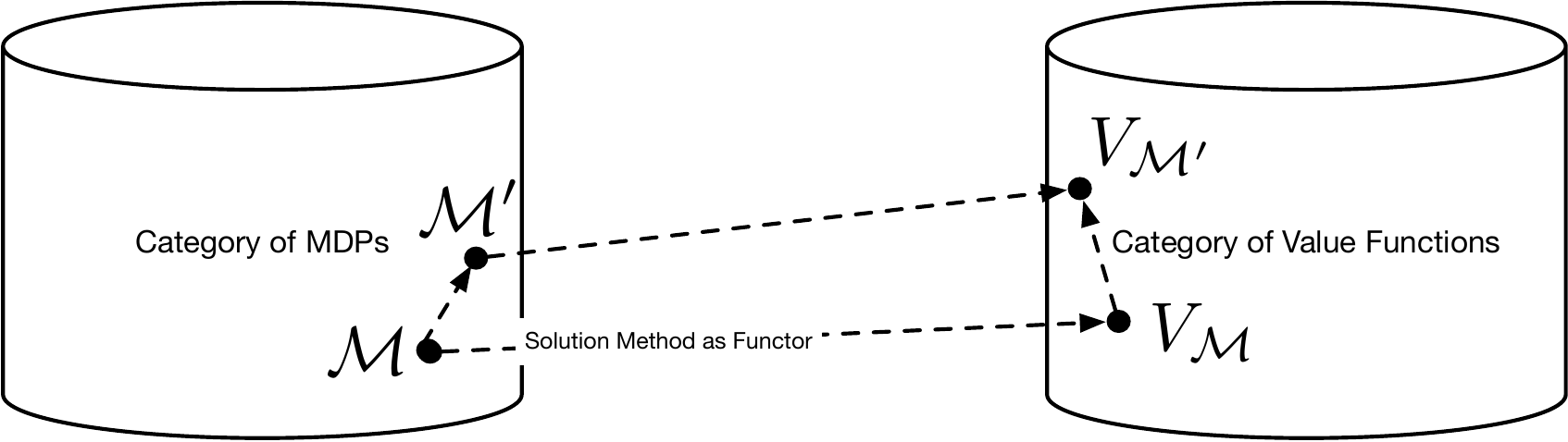}
    \caption{URL models MDP solution methods as {\em functors} mapping from the category of MDPs to the category of value functions. The ensemble of all such methods forms a functor category. Universal properties of these categories are studied in this paper. For example, the category of (action) value functions is shown to form a {\em topos} \citep{maclane:sheaves}.}
    \label{fig:mdpfunctor}
\end{figure}

Figure~\ref{fig:mdpfunctor} gives a high-level illustration of the first part of the paper. We conceptualize the solution of RL problems modeled as MDPs as {\em functors} that map from a {\em category} of MDPs ${\cal C}_{MDP}$ to a {\em category} of value functions ${\cal C}_V$. A functor $F: {\cal C}_{MDP} \rightarrow {\cal C}_V$ are specified by an {\em object} function mapping a particular MDP ${\cal M}$ to some associated value function $F {\cal M} = V_{\cal M}$, and $F {\cal M'} = V_{\cal M'}$, and an {\em arrow} function mapping some MDP homomorphism \citep{DBLP:conf/ijcai/RavindranB03} $f \in {\cal C}_{MDP}({\cal M}, {\cal M'})$ into a corresponding arrow $Ff \in {\cal C}_V(V_{\cal M}, V_{\cal M'})$. A similar structure exists for the category ${\cal C}_{PSR}$ of predictive state representations (PSRs) \citep{DBLP:conf/aaai/SoniS07}, whose objects are PSR models and whose arrows are PSR homomorphisms. 

Functors come in many varieties, but two important classes are used in this paper. 

\begin{definition}  
A {\bf {covariant functor}} $F: {\cal C} \rightarrow {\cal D}$ from category ${\cal C}$ to category ${\cal D}$, is defined as \mbox{the following: }

\begin{itemize} 
    \item An object $F X$ (also written as $F(X)$) of the category ${\cal D}$ for each object $X$ in category ${\cal C}$.
    \item An  arrow  $F(f): F X \rightarrow F Y$ in category ${\cal D}$ for every arrow  $f: X \rightarrow Y$ in category ${\cal C}$. 
   \item The preservation of identity and composition: $F \ id_X = id_{F X}$ and $(F f) (F g) = F(g \circ f)$ for any composable arrows $f: X \rightarrow Y, g: Y \rightarrow Z$. 
\end{itemize}
\end{definition}  

\begin{definition}  
A {\bf {contravariant functor}} $F: {\cal C} \rightarrow {\cal D}$ from category ${\cal C}$ to category ${\cal D}$ is defined exactly like the covariant functor, except all the arrows are reversed. 
\end{definition}  

We give a brief overview of RL in Section~\ref{rlintro} for readers unfamiliar with the field. Section~\ref{urlcat} illustrates the definition of categories over MDPs, using the familiar notions of MDP or PSR homomorphisms \citep{DBLP:conf/ijcai/RavindranB03,DBLP:conf/aaai/SoniS07}, although it is worth stressing that the arrows in an MDP or PSR category may represent any composable operation on these dynamical systems. Section~\ref{dpfuncat} begins a series of sections, illustrating how the space of solution methods in RL can be modeled as functor categories, as illustrated in Figure~\ref{fig:mdpfunctor}, first with exact solution methods, such as policy and value iteration, and linear programming. Section~\ref{rlfuncat} shows that simulation and RL-based methods can also be viewed in terms of functor categories. Section~\ref{deeprl} shows how deep RL can be modeled as a functor, building on the functorial characterization of backpropagation \citep{DBLP:conf/lics/FongST19}. 

\subsection{Part II: Generalizing RL to URL through Coalgebras} 

In the second part of the paper, we generalize RL to {\em universal RL} (URL). Section~\ref{introcoalgebra} introduces the topic of universal coalgebras, a categorical framework for dynamical systems, which generalizes a vast range of dynamical systems, from finite state automata, grammars and Turing machines, to stochastic dynamical systems like MDPs or PSRs by universal coalgebras \citep{jacobs:book,rutten2000universal,SOKOLOVA20115095}.  A coalgebra is simply defined as the structure 

\[ \alpha_F =  X \rightarrow F(X) \]

where $X$ is an object in some category ${\cal C}$, usually referred to as the {\em carrier}, and the functor $F$ defines the $F$-dynamics of the coalgebra. As an example of a coalgebra, consider the functor ${\cal P}_X: X \rightarrow 2^X$ that maps from a set $X$ to its powerset $2^X$ in the category ${\bf Sets}$ of sets. \citet{SOKOLOVA20115095} describes how to build an entire language of {\em probabilistic} coalgebras, using the context-free grammar: 

\[ F \coloneqq \textunderscore  \ | \ A \ | \ \textunderscore  ^A  \ | \ {\cal P} \ | \ {\cal D} \  | \ F \circ F \ | \ F \times F \ | \ F + F \]

where $F$ denotes a functor constructed from this grammar. Here, $\textunderscore$ is the identity functor over the category {\bf Sets}. $A$ is the constant functor mapping any set to the fixed set $A$. ${\bf id}^A$ defines a mapping from any set $X$ to the set of all functions from $A$ to $X$. The {\em powerset} functor ${\cal P}$ maps a set $X$ to its collection of subsets. Most importantly, the probability distribution functor ${\cal D}$ is defined as follows. 

\begin{definition} 
    The {\bf probability distribution functor} ${\cal D}$ is defined as ${\cal D}: {\bf Sets} \rightarrow {\bf Sets}$ maps a set $X$ to ${\cal D} X = \{ \mu: X \rightarrow \mathbb{R}^{\geq 0} | \mu[X] = 1 \}$, and a function $f: X \rightarrow Y$ to ${\cal D}f: {\cal D} X \rightarrow {\cal D} Y$ as $({\cal D} f)(\mu) = \lambda.\mu[f^{-1}(\{y \})$. 
\end{definition} 

In plain English, a distribution functor constructs a probability distribution over any set that has finite support, and given any function $f$ from set $X$ to set $Y$, maps any element $y$ in the codomain ${\cal D} Y$ to the probability mass assigned by to its preimage by ${\cal D}f$.   We can now define an entire family of stochastic coalgebras as shown in Table~\ref{stochcoalg}. We follow the terminology introduced in \citep{SOKOLOVA20115095}. To translate into the RL language, Segala systems correspond to MDPs, and Vardi systems are essentially concurrent Markov chains. It is possible to derive properties of the entire family of such coalgebras using the universal properties sketched out in the previous section, which we will refer the reader to \citep{SOKOLOVA20115095}. Our goal here is to adapt this categorical language into the URL framework, which we discuss next. 

\begin{table}[h]
    \centering
    \begin{tabular}{|c|c|c|} \hline 
         \bf{Coalg}$_F$ & $F$ & {\bf Explanation}  \\ \hline 
         {\bf MC} & {\cal D} & Markov chain \\ \hline 
        {\bf DLTS} & $(\textunderscore + 1)^A$ & Deterministic automata \\ \hline 
        {\bf LTS} & ${\cal P}(A \times \textunderscore ) \simeq {\cal P}^A $ & Non-deterministic automata  \\ \hline 
        {\bf React} &  $({\cal D} + 1)^A$ & Reactive systems \\ \hline 
        {\bf Generative} & ${\cal D}(A \times \textunderscore) + 1$ & Generative Systems \\ \hline 
        {\bf Str} &  ${\cal D} + (A \times \textunderscore) + 1$ & Stratified systems \\ \hline 
        {\bf Alt} & ${\cal D} + {\cal P}(A \times \textunderscore) $ & Alternating systems \\ \hline 
        {\bf Var} & ${\cal D}(A \times \textunderscore) + {\cal P}(A \times \textunderscore) $ & Vardi systems \\ \hline 
        {\bf SSeg} & ${\cal P}(A \times {\cal D}) $ & Simple Segala Systems \\ \hline 
        {\bf Seg} & ${\cal P} {\cal D}(A \times \textunderscore)$ & Segala systems \\ \hline 
        {\bf Bun} & ${\cal D} {\cal P}(A \times \textunderscore)$ & Bundle systems \\ \hline 
        {\bf PZ} & ${\cal P} {\cal D} {\cal P} (A \times \textunderscore)$ & Pneuli-Zuck systems \\ \hline 
        {\bf MG} & ${\cal P} {\cal D} {\cal P} (A \times \textunderscore \times \textunderscore)$ & Most general systems \\ \hline 
    \end{tabular}
    \caption{Stochastic Coalgebras in URL using the notation from  \citep{SOKOLOVA20115095}.}
    \label{stochcoalg}
\end{table}

\section{Asynchronous Distributed Minimization}

Before proceeding to outline our coinduction-based URL framework, we briefly review the induction-based framework of asynchronous stochastic approximation. Many RL algorithms, such as $Q$-learning, are essentially stochastic approximation methods \citep{kushner2003stochastic} that are amenable to parallel distributed computation techniques, and we briefly review this methodology here, particularly the work of \citet{bertsekas:pdc} that formed the basis for the convergence analysis of $Q$-learning by \citet{tsitsiklis}. It is, as we will see, couched in the language of {\em induction}. In sharp contrast, URL will be couched in the language of {\em coinduction}, which we will discuss in Section~\ref{introcoind} next.  In a nutshell, the classical problem studied in \citep{bertsekas:pdc} relates to solving the fixed point equation 

\[ F(x^*) = x^*, \ \ \ x^* \in \mathbb{R}^n\]

where the mapping $F$ is comprised of a set of component mappings $f_i$, which admit asynchronous parallel distributed computation.  In the language of coalgebras, this fixed point equation will be described as finding a {\em final coalgebra}, a problem that has been solved in great generality \citep{Aczel1988-ACZNS}. The space of solutions that $x^*$ lies in is assumed to be the Cartesian product 

\[ X = X_1 \times X_2 \ldots X_n \]

and the solutions are vectors of the form 

\[ x = (x_1, \ldots, x_n) \]

where the component functions $f_i$ assemble together as 

\[ F(x) = (f_1(x), \ldots, f_n(x)), \ \ \ \forall \ x \in X \]

and the fixed point of $F$ is computed using an asynchronous distributed version of the iterative method 

\[ x_i = f_i(x) ,  \ \ \ i=1, \ldots, n \]

Algorithm~\ref{adm} describes a formulation of asynchronous distributed minimization of a decomposable function, partly rephrased in terms of the coalgebraic language we will use in the URL framework. Later in the paper, we will give the coalgebraic URL generalization of this procedure as Algorithm~\ref{urlalgm}. 

\begin{algorithm}
\caption{Asynchronous Distributed Minimization \citep{bertsekas:pdc}} 
\label{adm}

{\bf Input:} Some function $F: X \rightarrow X$, where $X = \mathbb{R}^n$ or some (generalized) metric space decomposable as $X = X_1 \times X_2 \times \cdots, X_n$, and $F = \left( f_1(x), f_2(x), \ldots, f_n(x) \right)$, for all $x \in X$. 

{\bf Output:} A fixed point of $F$, namely an element $x^* \in X$ such that $F(x^*) = x^*$, which can be expressed also as $x_i^* = f_i(x^*)$, for all $i = 1, \ldots n$. This problem can be stated in the coalgebraic language of URL as finding a final coalgebra of a decomposable coalgebra. 

\begin{algorithmic}[1]
\REPEAT

\STATE At each time step $t \in T$, where $T = \{0, 1, \ldots, \}$, update some component $x_i$ of $x$ , using an asynchronous distributed version of the coalgebraic iteration: 

\[ x_i \rightarrow f_i(x) \]
\label{step1}

\STATE Each update  of $x_i$ is done in parallel by some ``processor" that may not have access to the latest values of all components of $x$, but we can assume that 

\[ x_i(t+1) = f_i(x1(\tau^1_i(t)), \ldots x_n(\tau^i_n(t))),  \ \ \forall t \in T^i\]

\STATE where $T^i$ is the set of time points where $x_i$ is updated, and $0 \leq \tau^i_j(t)  \leq t$, and at all times $t \notin T^i$, we assume that 

\[ x_i(t+1) = x_i(t) \]

\IF{the fixed point of $F$ is not reached}

\STATE Set $t = t+1$, and return to Step~\ref{step1}. 

\ELSE 

\STATE Set {\bf done} $\leftarrow$ {\bf true}. 

\ENDIF 

\UNTIL{{\bf done}}. 

\STATE  Return the fixed point $x^*$ of $F$. 

\end{algorithmic}
\end{algorithm} 

\subsection{Asynchronous Convergence Theorem} 
\label{act}

The general theory of convergence of such an approach to computing fixed points was later applied to the convergence of $Q$-learning by \citet{tsitsiklis}, as discussed below. The general convergence theorem in \cite{bertsekas:pdc} (Section 6.2) of Algorithm~\ref{act} proceeds by postulating that there is a sequence of nonempty sets $\{ X(k) \}$ such that

\[ \ldots \subset X(k+1) \subset X(k) \subset \cdots \subset X \]

satisfying the two conditions: 

\begin{enumerate}
    \item {\em Synchronous convergence condition}: Here, $f(x) \in X(k+1), \forall \ k \ \mbox{and} \ x \in X(k)$. Also, if $\{y^k \}$ is some sequence of vectors where $y^k \in X(k)$ for every $k$, then every limit point of $y^k$ is a fixed point of $F$. 

    \item {\em Box condition:} For every $k$, there exists a sets $X_i(k) \subset X_i$ such that 

    \[ X(k) = X_1(k) \times X_2(k) \times \cdots \times X_n(k) \]
\end{enumerate}

\citet{bertsekas:pdc} then prove the following {\em Asynchronous Convergence Theorem}: 

\begin{theorem}
\label{actthem}
    If the above two conditions hold, and the initial solution estimate 

    \[ x(0) = \left( x)1(0), \ldots, x_n(0) \right) \]

    belongs to the set $X(0)$, then every limit point of $\{x(t) \}$ is a fixed point of $F$. 
\end{theorem}

What's important here is that the actual proof given in \citep{bertsekas:pdc} is a proof based on {\em mathematical induction}: if the induction hypothesis is true for $k=0$, which it is since the initial estimate $x(0)$ is assumed to lie inside the set $X(0)$, then they show that there exists a time $t_{k+1}$ with the requisite properties.  URL sharply deviates from this framework in that it is based not on induction, and consequently {\em initial algebras} of well-founded sets, but on the principle of {\em coinduction} on {\em non-well-founded} sets. We will revisit the proof of the ACT theorem in Section~\ref{metricyoneda} using an interesting way to construct embeddings in generalized metric spaces. 

\subsection{Stochastic Approximation Theory for RL}

There are many ways to analyze the convergence of RL algorithms, from stochastic approximation using ordinary differential equations (ODEs) \citep{kushner2003stochastic,borkar} to parallel distributed computation \citep{tsitsiklis,bertsekas:pdc}.   As a concrete example, \citet{tsitsiklis} analyzes the convergence of the following iteration: 

\begin{equation}
\label{stoch-eqn}
    x_i(t+1) = x_i(t) + \alpha_i(t) \{ F_i(x^i(t)) - x_i(t) + w_i(t) \}, \ \ t \in T^i 
\end{equation}

where $F: \mathbb{R}^n \rightarrow \mathbb{R}^n$ is a mapping from $n$-dimensional Euclidean space back to itself, whose fixed point $F(x) = x$ is to be determined at some point $x^*$. This is essentially a coalgebraic equation, and the fixed point $F(x^*) = x^*$ is the final coalgebra. To find the fixed point of the mapping $F$ defined in Equation~\ref{stoch-eqn}, \cite{tsitsiklis} defines a sequence of nested $\Sigma$-algebras, denoted ${\cal F(t)}p^\infty_0$, on a probability space $(\Omega, {\cal F}, {\cal P})$ over which all the random variables $(x(t), \tau^i_j(t),\alpha_i(t), w_i(t))$, are ${\cal F}(t)$ measurable.

To guarantee convergence of stochastic approximation in RL, \citet{tsitsiklis} proposed a framework wherein RL algorithms could be solved using parallel distributed computation comprising of a series of decision makers, each of which updates a particular state-action pair $(s,a)$ asynchronously, and in parallel with no coordination between them. This framework builds on previous work described in \citet{bertsekas:pdc}. A key assumption made in this work, and applied to the convergence of $Q$-learning is that the resulting $\Sigma$-algebras ${\cal F}(t)$ are all measurable, meaning that the random variables in Equation~\ref{stoch-eqn} are all measurable at each time instant. Later,  we review and generalize the work of \citet{witsenhausen:1975} on the {\em intrinsic} model, a parallel distributed computational model, which addresses the problem of how to structure a sequence of information fields such that the resulting $\Sigma$-algebras are indeed measurable. It is this specific question that is somewhat left unaddressed in the work of \citet{tsitsiklis} and \citet{bertsekas:pdc}. We describe a categorical generalization of the Witsenhausen's framework, first proposed in our previous work on universal decision models (UDMs) \citep{sm:udm}, and here specialized to a category of coalgebras for URL.

\subsection{Using Diagrams to Specify URL Architectures}

A general theme in URL is the use of diagrams $F: {\cal J} \rightarrow {\cal C}$, where ${\cal J}$ is a finite category, and $F$ is a functor that maps each object $j \in {\cal J}$ to an object ${\cal Fj} \in {\cal C}$ and each arrow $f: j \rightarrow j'$ in ${\cal J}$ to the corresponding arrow $Ff: Fj \rightarrow Fj'$ in ${\cal C}$. Let us relate this abstract view to the structure defined in Algorithm~\ref{adm}. 

Algorithm~\ref{adm} uses a simple product structure, which corresponds to a diagram of the form ${\cal J}  = \bullet \bullet \cdots \bullet$, where each $\bullet$ is a placeholder for an actual object in category ${\cal C}$, which in Algorithm~\ref{adm} corresponds to one whose objects are defined as elements of $\mathbb{R}^n$, and whose arrows are mappings over $\mathbb{R}^n$. The actual product structure itself emerges as the {\em limit} of the diagram $F: {\cal J} \rightarrow {\cal C}$. As we will see in later sections of this paper, we can conceptualize all of the previous work in RL in terms of diagrams, ranging from simple table-lookup methods, which corresponds to a discrete category with only identity arrows, to deep RL where the diagram is specified by a neural network as shown in \citep{DBLP:conf/lics/FongST19}. We will generalize Algorithm~\ref{adm} eventually to Algorithm~\ref{urlalgm}, where the product structure will be replaced by a functor diagram over a category of coalgebras. 

\section{What are Universal Coalgebras?}
\label{introcoalgebra}

Rather than deal with concrete models, such as MDPs or PSRs, URL is formulated in terms of universal coalgebras \citep{rutten2000universal,jacobs:book}, a categorical language for specifying dynamical systems. At a high level, algebras {\em combine} entities, like $1+1 = 2$. On the other hand, coalgebras {\em expand} entities, and are invaluable in describing the behavior of automata, grammars, and programs that generate infinite data streams. A  standard example that is often used to illustrate coalgebras, and provides a foundation for many AI and  ML applications, is that of a {\em labeled transition system}. 

\begin{definition} 
    A {\bf labeled transition system} (LTS) $(S, \rightarrow_S, A)$ is defined by a set $S$ of states, a transition relation $\rightarrow_S \subseteq S \times A \times S$, and a set $A$ of labels (or equivalently, ``inputs" or ``actions"). We can define the transition from state $s$ to $s'$ under input $a$ by the transition diagram $s \xrightarrow[]{a} s'$, which is equivalent to writing $\langle s, a,  s' \rangle \in \rightarrow_S$. The ${\cal F}$-coalgebra for an LTS is defined by the functor 

    \[ {\cal F}(X) = {\cal P}(A \times X) = \{V | V \subseteq A \times X\} \]
\end{definition} 
We can also define a category of $F$-coalgebras over any category ${\cal C}$, where each object is a coalgebra, and the morphism between two coalgebras is defined as follows, where $f: A \rightarrow B$ is any morphism in the category ${\cal C}$. 
\begin{definition} 
Let $F: {\cal C} \rightarrow {\cal C}$ be an endofunctor. A {\em homomorphism} of $F$-coalgebras $(A, \alpha)$ and $(B, \beta)$ is an arrow $f: A \rightarrow B$ in the category ${\cal C}$ such that the following diagram commutes:

\begin{center}
\begin{tikzcd}
  A \arrow[r, "f"] \arrow[d, "\alpha"]
    & B \arrow[d, "\beta" ] \\
  F(A) \arrow[r,  "F(f)"]
& F(B)
\end{tikzcd}
\end{center}
\end{definition} 

For example, consider two labeled transition systems $(S, A, \rightarrow_S)$ and $(T, A, \rightarrow_T)$ over the same input set $A$, which are defined by the coalgebras $(S, \alpha_S)$ and $(T, \alpha_T)$, respectively. An $F$-homomorphism $f: (S, \alpha_S) \rightarrow (T, \alpha_T)$ is a function $f: S \rightarrow T$ such that $F(f) \circ \alpha_S  = \alpha_T \circ f$. Intuitively, the meaning of a homomorphism between two labeled transition systems means that: 
\begin{itemize}
    \item For all $s' \in S$, for any transition $s \xrightarrow[]{a}_S s'$ in the first system $(S, \alpha_S)$, there must be a corresponding transition in the second system $f(s) \xrightarrow[]{a}_T f(s;)$ in the second system. 

    \item Conversely, for all $t \in T$, for any transition $t \xrightarrow[]{a}_T t'$ in the second system, there exists two states $s, s' \in S$ such that $f(s) = t, f(t) = t'$ such that $s \xrightarrow[]{a}_S s'$ in the first system. 
\end{itemize}
If we have an $F$-homomorphism $f: S \rightarrow T$ with an inverse $f^{-1}: T \rightarrow S$ that is also a $F$-homomorphism, then the two systems $S \simeq T$ are isomorphic. If the mapping $f$ is {\em injective}, we have a  {\em monomorphism}. Finally, if the mapping $f$ is a surjection, we have an {\em epimorphism}. 
In the RL literature, there has been substantial interest in MDP ``bisimulation" \citep{DBLP:conf/ijcai/RavindranB03,DBLP:conf/aaai/CastroP10,DBLP:conf/aaai/RuanCPP15}, all of which are special cases of the more general notion of  {\em bisimulation} in universal coalgebras. Intuitively, bisimulation allows us to construct a more ``abstract" representation of a dynamical system that is still faithful to the original system. We will explore many applications of the concept of bisimulation to AI and ML systems in this paper. We introduce the concept in its general setting first, and then in the next section, we will delve into concrete examples of bisimulations. 

\begin{definition} 
Let $(S, \alpha_S)$ and $(T, \alpha_T)$ be two systems specified as coalgebras acting on the same category ${\cal C}$. Formally, a $F$-{\bf bisimulation} for coalgebras defined on a set-valued functor $F: {\bf Set} \rightarrow {\bf Set}$ is a relation $R \subset S \times T$ of the Cartesian product of $S$ and $T$ is a mapping $\alpha_R: R \rightarrow F(R)$ such that the projections of $R$ to $S$ and $T$ form valid $F$-homomorphisms.

\begin{center}
\begin{tikzcd}
  R \arrow[r, "\pi_1"] \arrow[d, "\alpha_R"]
    & S \arrow[d, "\alpha_S" ] \\
  F(R) \arrow[r,  "F(\pi_1)"]
& F(S)
\end{tikzcd}
\end{center}

\begin{center}
\begin{tikzcd}
  R \arrow[r, "\pi_2"] \arrow[d, "\alpha_R"]
    & T \arrow[d, "\alpha_T" ] \\
  F(R) \arrow[r,  "F(\pi_2)"]
& F(T)
\end{tikzcd}
\end{center}

Here, $\pi_1$ and $\pi_2$ are projections of the relation $R$ onto $S$ and $T$, respectively. Note the relationships in the two commutative diagrams should hold simultaneously, so that we get 

\begin{eqnarray*}
    F(\pi_1) \circ \alpha_R &=& \alpha_S \circ \pi_1 \\
    F(\pi_2) \circ \alpha_R &=& \alpha_T \circ \pi_2 
\end{eqnarray*}

Intuitively, these properties imply that we can ``run" the joint system $R$ for one step, and then project onto the component systems, which gives us the same effect as if we first project the joint system onto each component system, and then run the component systems. More concretely, for two labeled transition systems that were considered above as an example of an $F$-homomorphism, an $F$-bisimulation between $(S, \alpha_S)$ and $(T, \alpha_T)$ means that  there exists a relation $R \subset S \times T$ that satisfies for all $\langle s, t \rangle \in R$

\begin{itemize}
    \item For all $s' \in S$, for any transition $s \xrightarrow[]{a}_S s'$ in the first system $(S, \alpha_S)$, there must be a corresponding transition in the second system $f(s) \xrightarrow[]{a}_T f(s;)$ in the second system, so that $\langle s', t' \rangle \in R$

    \item Conversely, for all $t \in T$, for any transition $t \xrightarrow[]{a}_T t'$ in the second system, there exists two states $s, s' \in S$ such that $f(s) = t, f(t) = t'$ such that $s \xrightarrow[]{a}_S s'$ in the first system, and $\langle s', t' \rangle \in R$.
\end{itemize}

\end{definition}

There are a number of basic properties about bisimulations, which we will not prove, but are useful to summarize here: 

\begin{itemize}
    \item If $(R, \alpha_R)$ is a bisimulation between systems $S$ and $T$, the inverse $R^{-1}$ of $R$ is a bisimulation between systems $T$ and $S$. 

    \item Two homomorphisms $f: T \rightarrow S$ and $g:T \rightarrow U$ with a common domain $T$ define a {\em span}. The {\em image} of the span $\langle f, g \rangle(T) = \{ \langle f(t), g(t) \rangle | t \in T \}$ of $f$ and $g$ is also a bisimulation between $S$ and $U$. 

    \item The composition $R \circ Q$ of two bisimulations $R \subseteq S \times T$ and $Q \subseteq T \times U$ is a bisimulation between $S$ and $U$. 

    \item The union $\cup_k R_k$ of a family of bisimulations between $S$ and $T$ is also a bisimulation. 

    \item The set of all bisimulations between systems $S$ and $T$ is a complete lattice, with least upper bounds and greatest lower bounds given by: 

    \[ \bigvee_k R_k = \bigcup_k R_k \]

    \[ \bigwedge_K R_k = \bigcup \{ R | R \ \mbox{is a bisimulation between} S \ \mbox{and} \ T \ \mbox{and} \ R \subseteq \cap_k R_k \} \]

    \item The kernel $K(f) = \{ \langle s, s' \rangle | f(s) = f(s') \}$ of a homomorphism $f: S \rightarrow T$ is a bisimulation equivalence.

    \end{itemize}

\subsection{MDPs as Coalgebras}

\citet{feys} and \citet{kozen} have previously explored coalgebraic formulations of MDPs and dynamic programming. The main difference with URL is our emphasis on solving universal coalgebras -- meaning to find the final coalgebra -- by stochastic sampling methods, rather than assuming the $F$-dynamics is fully known, as in their treatments. We refer the reader to their original papers for details of their particular approaches. We will review the basics of MDPs and RL in Section~\ref{rlintro}. 

\section{What is Coinduction?}
\label{introcoind} 

As the term {\em coinduction} may be unfamiliar to many readers of this paper, we give a brief review of the literature on this topic \citep{rutten2000universal}. The earliest work in machine learning was the problem of {\em identification in the limit},  first studied by Gold \citep{GOLD1967447} and Solomonoff \citep{SOLOMONOFF19641}, and subsequently extended to the stochastic setting by Valiant \cite{DBLP:journals/cacm/Valiant84} and Vapnik \cite{DBLP:journals/tnn/Vapnik99}. Inductive inference is related to mathematical induction, which in the categorical setting, refers to {\em initial algebras} over {\em well-founded sets}.  In contrast, {\em coinductive inference} arose originally in the study of non-well-founded sets \cite{Aczel1988-ACZNS},  and subsequently in universal coalgebras \cite{jacobs:book,rutten2000universal}. 

\begin{figure}[t]
      % Give a unique label
\includegraphics[scale=.5]{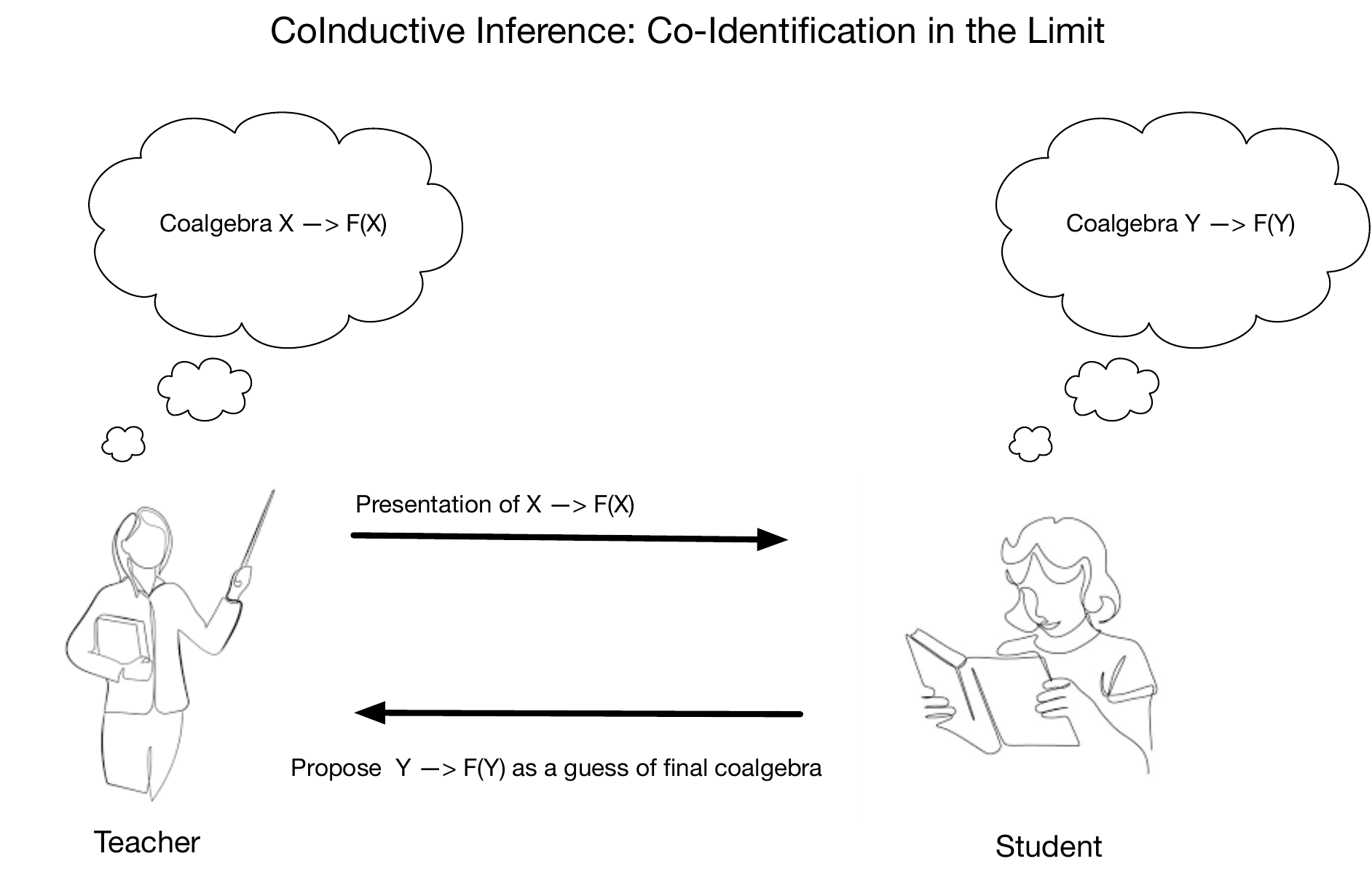}
%\picplace{5cm}{2cm} % Give the correct figure height and width in cm
%
\caption{The coinductive inference framework for URL proposed in this paper models the process of machine learning as discovering a final coalgebra in the category of coalgebras. Rather than enumerating sets, as in inductive inference, representations of non-well-founded sets as accessible pointed graphs (APGs) or universal coalgebras are enumerated. Coidentification refers to the use of {\em bisimulation}, a fundamental relation used to compare two non-well-founded sets or universal coalgebras.}
\label{coindfig} 
\end{figure}

Coinductive inference provides a novel theoretical foundation for ML that is based on the conceptualization of non-well-founded sets in terms of accessible pointed graphs (APGs) \cite{Aczel1988-ACZNS}, a representation that uses graphs not sets. The original motivation was how to deal with infinite data streams or data structures in computer science.   The term coinductive inference is based on {\em coinduction},  a proof principle, developed initially by Peter Aczel in his ground-breaking work on non-well-founded circular (or recursively defined) sets \cite{Aczel1988-ACZNS}. Figure~\ref{coindfig} illustrates the general framework of coinductive inference that is based on the theory of universal coalgebras. Comparing the framework for inductive inference presented above, the major difference is the use of AGPs or universal coalgebras as the formalism for describing the language used by the teacher to specify a particular model from which data is generated, and by the learner to produce a solution. Universal coalgebras provide a universal language for AI and ML in being able to describe a wide class of problems, ranging from causal inference, (non)deterministic and probabilistic finite state machines, game theory and reinforcement learning. 

To take some concrete examples, let us begin with the standard problem of learning a deterministic finite state machine model. We can specify a finite state machine model as a {\em labeled transition system} $(S, \rightarrow_S, A)$ consisting of a set $S$ of states, a transition relation $\rightarrow_S \subseteq S \times A \times S$, where $A$ is a set of input labels. Let us define a functor ${\cal B}(X) = {\cal P}(A \times X) = \{V | V \subseteq A \times X$ for any set $X$. ${\cal 
B}$ is then an endofunctor on the category of sets, which specifies the transition dynamics of the finite state machine. We can then represent a labelled transition system as a universal coalgebra of the following type: 

\[ \alpha_S: S \rightarrow {\cal B}(S), \ \ \ \ s \rightarrow \{(a, s') | s \xrightarrow[]{a} s' \} \]

for some fixed set of input symbols $A$. The coinductive inference framework assumes the teacher selects a particular coalgebra specifying a finite state machine and generates a presentation of it for the learner. Upon receiving a sequence of examples, the learner produces a hypothesis coalgebra describing the finite state machine. The learner's ultimate goal can be succinctly summarized as discovering the final coalgebra that represents the minimal finite state machine model that is isomorphic to the coalgebra selected by the teacher. 

\subsection{Dynamic Programming and Metric Coinduction} 
\label{metriccoinduction}

It is commonplace in ML and RL to investigate successive approximation algorithms in {\em metric spaces}, where some notion of distance is available. A standard procedure for checking the convergence of dynamic programming and RL methods is to show that the Bellman operator $T(V)$ is a {\em contraction} in some norm in Euclidean space \citep{DBLP:books/lib/Bertsekas05}. We discuss here the principle of {\em metric coinduction} \citep{kozen}, which introduced the idea of coinduction in metric spaces, as well as applied it to show the convergence of dynamic programming methods for solving MDPs. We will see later in Section~\ref{metricyoneda} how to generalize this framework using the metric Yoneda Lemma \citep{BONSANGUE19981} to generalized metric spaces. 

Let us consider a metric space $(V, d)$, and a function $F: V \rightarrow V$ that is {\em contractive} in that for some $0 \leq c < 1$ such that for all $u, v \in V$, 

\[ d(F(u), F(v)) \leq c \cdot d(u,v) \]

A continuous function $F$ is {\em eventually contractive} if $F^n$ is contractive for some $n \geq 1$. Fixed points of contractive maps $F$ can be found as the limit of a Cauchy sequence 

\[ u, F(u), F^2(u), \cdots, \]

with a standard argument being that 

\[ d(F^{n+m}(u), F^n(u)) \leq c^n (1 - c^m) (1 - c)^{-1} \cdot d(F(u), u) \]

\citet{kozen} introduce the following metric coinduction rule: 

\begin{definition} \citep{kozen}
    {\bf The Coinduction Rule}: If $\phi$ is a closed nonempty subset of a metric space $V$, and if $F$ is an eventually contractive map on $V$ that preserves $\phi$, then the unique fixed point $u^*$ of $F$ is in $\phi$. 
\end{definition}

To understand further why this rule is an example of coinduction, one has to show a category of coalgebras and that this rule essentially equivalent to saying that a certain coalgebra is {\em final} in this category. An object in a category is {\em final} if there exists a unique arrow {\em into} it from every other object in the category. \citet{kozen} show that this property is indeed a consequence of such a category of coalgebras, and we refer the reader to their paper for details. \citet{kozen} apply the metric coinduction principle to many interesting problems, including the property that any irreducible and aperiodic Markov chain converges to the stationary distribution, a rather intricate proof in the usual inductive setting, but much simpler in the coinductive setting. They also show that in MDPs, stochastic policies do not perform better than deterministic strategies. However, \citet{kozen} do not use coinduction to analyze the behavior of RL methods, which is the main topic of our paper. We can use metric coinduction to analyze the convergence of Algorithm~\ref{adm}, where the key difference is that there is no induction step! The box condition and the synchronous convergence condition are sufficient to show that the induced mapping in Algorithm~\ref{adm} is eventually contractive in some (generalized) metric space. 

It is also worth pointing out that the fundamental notion of a Cauchy sequence and a complete metric space are actually consequence of applying the famous Yoneda Lemma \citep{maclane:71} in a generalized metric space \citep{BONSANGUE19981}. We refer the reader to this paper for details, which are also discussed in our earlier work on universal imitation games \citep{sm:uig}.  We will return to this topic in Section~\ref{metricyoneda}. 

\subsection{Final Coalgebra Theorem} 

There is a large and rich literature on conditions under which a category of coalgebras has a final coalgebra. \citet{Aczel1988-ACZNS} discusses this problem in the context of non-well-founded sets, which he models using a graph representation. \citet{aczel-final-coalgebra-thm} proves that every set-based functor on the category of classes has a final coalgebra. The key idea in coinduction is that one can exploits such general results to simplify the otherwise tedious proofs that are required to show convergence of successive approximation methods. \citet{rutten:streams} analyzes the behavior of infinite data streams using coinduction. A wealth of additional examples are given in the book by \citet{jacobs:book} on coalgebra. 

\section{Introduction to RL}
\label{rlintro}

Most readers will not need a review of basic RL (e.g., see \citep{bertsekas:rlbook} for a more detailed treatment), but we include a basic introduction to fix our notation. We will not get into more sophisticated proximal RL methods, such as in our previous work \citep{mahadevan2014proximal}, or methods for learning representation and control, such as our work on proto-value functions \citep{DBLP:conf/icml/Mahadevan05}. 

The learning environment for sequential decision-making is generally modeled
by the well-known \textbf{\emph{Markov Decision Process}}\citep{PUTERMAN1990331} $M=(S,A,P,R,\gamma)$,
which is derived from a Markov chain.
\begin{definition} 
(Markov Chain)\label{def:mc}: A \emph{Markov Chain} is a stochastic process defined as $M=(S,P)$. At each time step $t=1,2,3,\cdots$,
the agent is in a state ${s_{t}}\in S$, and the state transition
probability is given by the state transition kernel $P:S\times S\to\mathbb{{R}}$
satisfying $||P|{|_{\infty}}=1$, where $P({s_{t}}|{s_{t-1}})$ is
the state-transition probability from state $s_{t-1}$ at time step
$t-1$ to the state $s_{t}$ at time step $s_{t}$.
\end{definition} 

As we will see in Section~\ref{stochcoalg}, Markov chains are a simple type of stochastic coalgebra defined by a functor ${\cal D}$, the distribution functor \citep{SOKOLOVA20115095}. A Markov decision process (MDPs) is comprised of a set of states $S$,
a set of (possibly state-dependent) actions $A$ ($A_{s}$), a dynamical
system model comprised of the transition probabilities $P_{ss'}^{a}$
specifying the probability of transition to state $s'$ from state
$s$ under action $a$, and a reward model $R$.
\begin{definition} 
(Markov Decision Process)\label{def:mdp}\citep{PUTERMAN1990331}: A Markov
Decision Process is a tuple $(S,A,P,R,\gamma)$ where $S$ is a finite
set of states, $A$ is a finite set of actions, $P:S\times A\times S\to[0,1]$
is the transition kernel, where $P(s,a,s')$ is the probability of
transmission from state $s$ to state $s'$ given action $a$, and
reward $r:S\times A\to{\mathbb{R}{}^{+}}$ is a reward function (or equivalently, a cost function), $0\leq\gamma<1$
is a discount factor.
\end{definition}

\subsection{Basics of Reinforcement Learning}

A policy $\pi:S\rightarrow A$ is a deterministic (stochastic) mapping
from states to actions.
\begin{definition} 
\noindent\label{def:pol}(Policy): A deterministic stationary policy
$\pi:S\to A$ assigns an action to each state of the Markov decision
process. A stochastic policy $\pi:S\times A\to[0,1]$.
%The
%set of all stochastic stationary policies is denoted as $\Pi$. Given
%$\Pi$ and a $\sigma$-algebra $F$ defined on $\Pi$, $(\Pi,F)$
%constitutes a topological space called policy space.
\end{definition} 
Value functions are used to compare and evaluate the performance of
policies.
\begin{definition} 
\noindent\label{def:value}(Value Function): A value function w.r.t
a policy $\pi$ termed as ${V^{\pi}}:S\to\mathbb{R}$ assigns each state the
\emph{expected sum of discounted} rewards (or equivalently, minimize the discounted sum of costs) 
\end{definition} 
\begin{equation}
V^{\pi}=\mathbb{E}\left[{\sum\limits _{i=1}^{t}{{\gamma^{i-1}}{r_{i}}}}\right]
\end{equation}

The goal of reinforcement learning is to find a (near-optimal) policy
that maximizes the value function. $V^{\pi}$ is a fixed-point of
the Bellman equation: 
\begin{equation}
{V^{\pi}}({s_{t}})=\mathbb{E}\left[{r({s_{t}},{\pi(s_{t})})+\gamma{V^{\pi}}({s_{t+1}})}\right]\label{eq:bellman_eq}
\end{equation}

As we will see in URL, this fixed point can be seen as the final coalgebra of a category of coalgebras defined by the MDP.  Equation (\ref{eq:bellman_eq}) can be written in a concise form by introducing the Bellman operator $T^{\pi}$ w.r.t a policy $\pi$
and denoting the reward vector as $R^{\pi}\in{\mathbb{R}^{n}}$ where ${R^{\pi}_{i}}=\mathbb{E}[r({s_{i}},\pi(s_{i}))]$.

\begin{equation}
V^{\pi}=T^{\pi}(V^{\pi})=R^{\pi}+\gamma P^{\pi}V^{\pi}\label{eq:bellman_op}
\end{equation}

Any optimal policy $\pi^{*}$ defines the unique optimal value function
$V^{*}$ that satisfies the nonlinear system of equations:
\begin{equation}
V^{^{*}}(s)=\max_{a}\sum_{s'}P_{ss'}^{a}\left(R_{ss'}^{a}+\gamma V^{*}(s')\right)\label{eq:Vopt}
\end{equation}

\subsection{Action Value Function Approximation}

The most popular and widely used RL method is temporal difference
(TD) learning \citep{DBLP:books/lib/SuttonB98}. TD learning is a stochastic approximation
approach to solving Equation (\ref{eq:Vopt}). The {\em state-action
value} $Q^{*}(s,a)$ represents a convenient reformulation of the
value function, defined as the long-term value of performing $a$
first, and then acting optimally according to $V^{*}$:
\begin{equation}
Q^{*}(s,a)=\mathbb{E}\left(r_{t+1}+\gamma\max_{a'}Q^{*}(s_{t+1},a')|s_{t}=s,a_{t}=a\right)\label{qvalues}
\end{equation}
where $r_{t+1}$ is the actual reward received at the next time step,
and $s_{t+1}$ is the state resulting from executing action $a$ in
state $s_{t}$. The (optimal) action value formulation is convenient
because it can be approximately solved by a temporal-difference (TD)
learning technique called Q-learning \cite{DBLP:books/lib/SuttonB98}. The simplest
TD method, called TD($0$), estimates the value function associated
with the fixed policy using a normal stochastic gradient iteration,
where $\delta_{t}$ is called temporal difference error:
\begin{equation}
\begin{array}{l}
{V_{t+1}}({s_{t}})={V_{t}}({s_{t}})+{\alpha_{t}}{\delta_{t}}\\
{\delta_{t}}={r_{t}}+\gamma{V_{t}}({s_{t+1}})-{V_{t}}({s_{t}})
\end{array}
\end{equation}

TD($0$) converges to the optimal value function $V^{\pi}$ for policy
$\pi$ as long as the samples are ``on-policy'', namely following
the stochastic Markov chain associated with the policy; and the learning
rate $\alpha_{t}$ is decayed according to the Robbins-Monro conditions
in stochastic approximation theory: $\sum_{t}\alpha_{t}=\infty,\sum_{t}\alpha_{t}^{2}<\infty$
\cite{borkar}. When the set of states $S$ is large, it is often
necessary to approximate the value function $V$ using a set of handcrafted
basis functions (e.g., polynomials, radial basis functions, wavelets
etc.) or automatically generated basis functions \cite{mahadevan2014proximal}.
In linear value function approximation, the value function is assumed
to lie in the linear spanning space of the basis function matrix $\Phi$
of dimension $|S|\times d$, where it is assumed that $d\ll|S|$.
Hence,
\begin{equation}
V^{\pi}\approx V_{\theta}=\Phi\theta
\end{equation}
The equivalent TD($0$) algorithm for linear function approximated
value functions is given as:
\begin{equation}
\begin{array}{l}
{\theta _{t + 1}} = {\theta _t} + {\alpha _t}{\delta _t}\phi ({s_t})\\
{\delta _t} = {r_t} + \gamma \phi {({s_{t + 1}})^T}{\theta _t} - \phi {({s_t})^T}{\theta _t}
\end{array}
\label{td0}
\end{equation}

\subsection{Proximal RL} 

In a previous paper \citep{mahadevan2014proximal}, we proposed a comprehensive theory of {\em proximal RL} in primal-dual spaces. We briefly summarize this work as an example of a more sophisticated framework that could be extended to URL, but which we will leave aside for now. Proximal RL addresses the following challenges: (i) how to design reliable, convergent, and robust reinforcement learning algorithms (ii) how to guarantee that reinforcement learning satisfies pre-specified ``safely" guarantees, and remains in a stable region of the parameter space (iii) how to design ``off-policy" temporal difference learning algorithms in a reliable and stable manner, and finally (iv) how to integrate the study of reinforcement learning into the rich theory of stochastic optimization. The  most important idea that proximal RL uses  is the notion of {\em primal dual spaces} connected through the use of a {\em Legendre} transform. This allows temporal difference updates to occur in dual spaces, allowing a variety of important technical advantages. The Legendre transform, elegantly generalizes past algorithms for solving reinforcement learning problems, such as {\em natural gradient} methods, which we show relate closely to the previously unconnected framework of {\em mirror descent} \citep{beck:phd}.  Equally importantly, proximal operator theory enables the systematic development of {\em operator splitting} methods that show how to safely and reliably decompose complex products of gradients that occur in recent variants of gradient-based temporal difference learning. This key technical innovation makes it possible to finally design ``true" stochastic gradient methods for reinforcement learning. Finally, Legendre transforms enable a variety of other benefits, including modeling sparsity and domain geometry. 

\subsection{Categorical Function Approximation} 

Category theory provides powerful tools for approximating functions on categories, including ideas from algebraic topology \citep{may1999concise}. We have explored these ideas in past work in categorical causality, such as the use of simplicial sets that construct ``nice" CW-complexes in topology \citep{DBLP:journals/entropy/Mahadevan23}. As we said earlier in Section~\ref{sec:intro}, we can construct general architectures in URL in terms of functor diagrams $F: {\cal J} \rightarrow {\cal C}$, where ${\cal J}$ is a finite category that specifies some set of abstract objects and their relationship that gets mapped into the actual RL category. As we will see in Section~\ref{deeprl}, deep RL itself corresponds precisely to such a functor diagram. 

\section{Categories for URL} 
\label{urlcat}
\begin{figure}[t]
    \centering
    \includegraphics[width=0.25\linewidth]{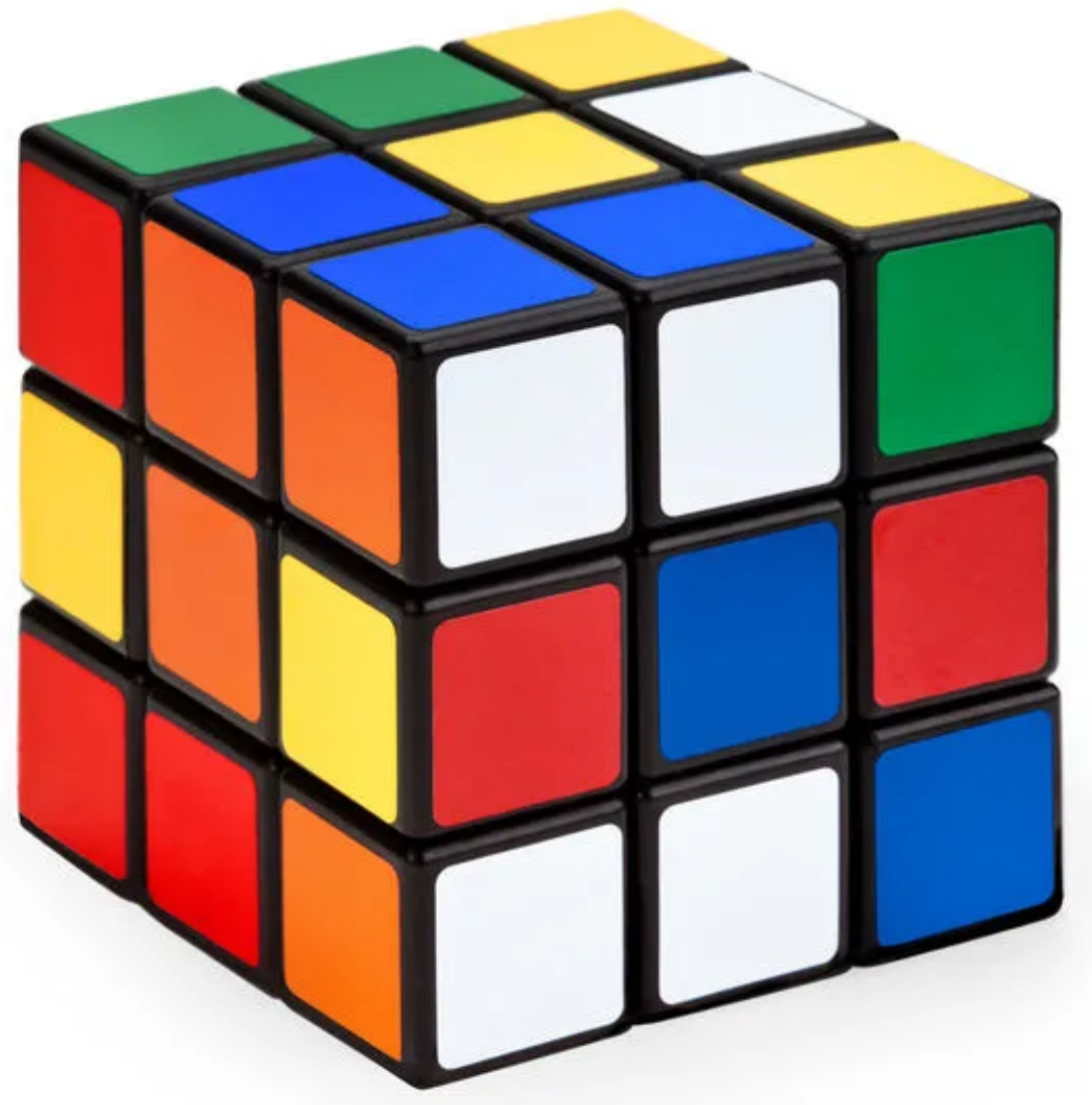} \hspace{0.5in}
      \includegraphics[width=0.25\linewidth]{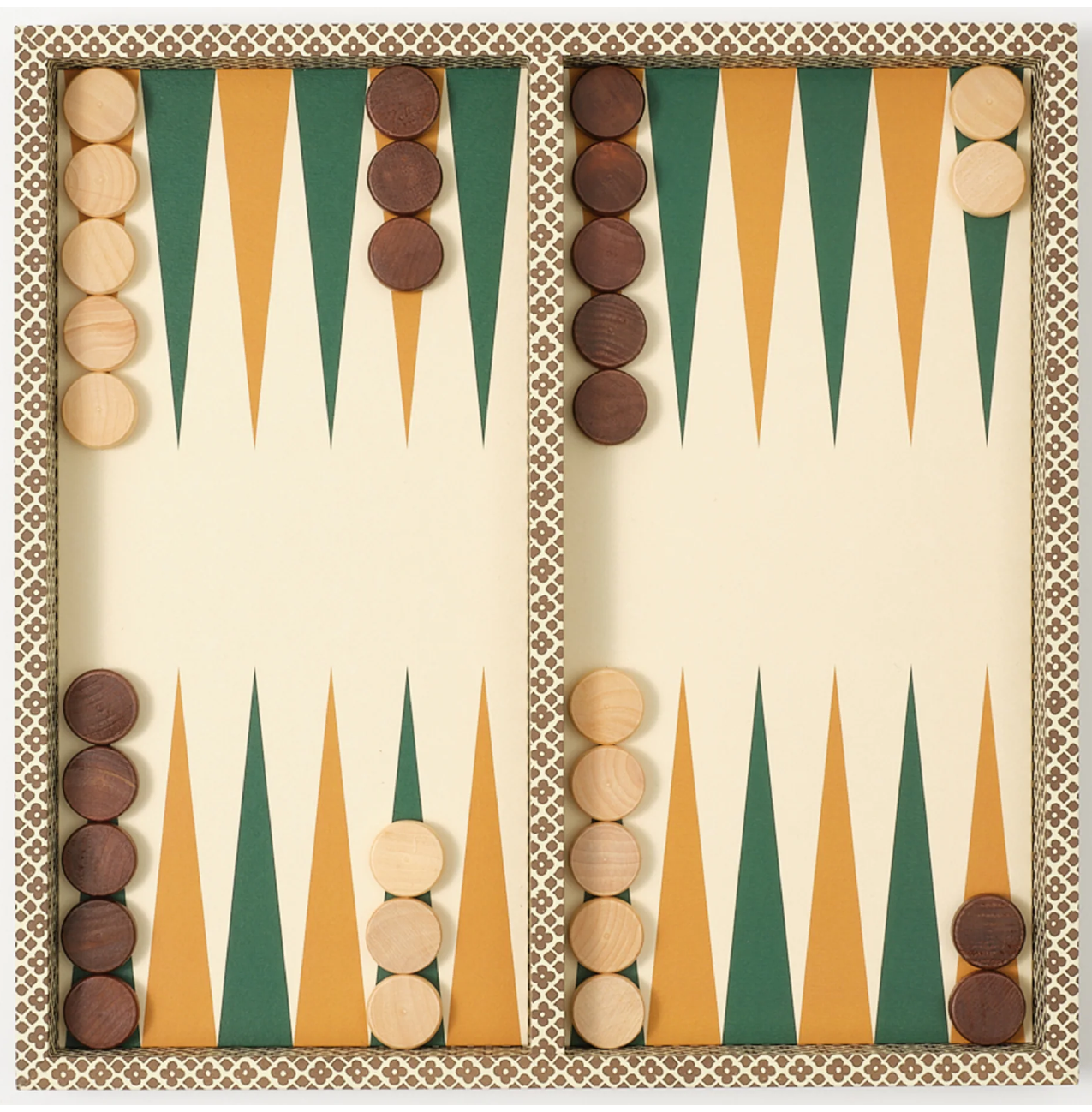}
    \caption{Two popular games easily solvable by RL can be modeled as categories.}
    \label{fig:rl-examples}
\end{figure}
To warm up before the main course begins, let us consider simple examples of how to define categories for (U)RL. Formally, a {\em category} ${\cal C}$ is a collection of {\em objects} $\mbox{ob}({\cal C})$ and a collection of {\em arrows} $\mbox{arr}({\cal C})$. Between each pair of objects $c,c' \in {\cal C}$, we are given a collection of arrows ${\cal C}(c, c')$. Arrows compose in an intuitive way: the arrow $f: c \rightarrow c'$ composes with the arrow $g: c' \rightarrow d$, which is written as $g \circ f: c \rightarrow d$. There is a privileged {\em identity} arrow associated with each object $\mbox{id}_c: c \rightarrow c$. 

Figure~\ref{fig:rl-examples} gives two examples of popular games easily solvable by RL, which can be modeled as categories. The Rubiks cube is an example of a category called a {\em groupoid} \citep{may1999concise}: in such a category, every arrow is invertible. All the moves in Rubiks cube are of course invertible. Backgammon, one of the earliest successes of deep RL \citep{TESAURO2002181}, can be modeled under a fixed (say random) policy as an absorbing Markov chain, and as such defines a stochastic coalgebra \citep{SOKOLOVA20115095}. 

A {\em functor} $F: {\cal C} \rightarrow {\cal D}$ consists of two functions: the {\em object} function  maps each object $c \in {\cal C}$ to the corresponding $F c \in {\cal D}$; the {\em arrow} function maps each arrow $f \in \mbox{arr}({\cal C})$ to the corresponding arrow $F f \in \mbox{ar}({\cal D})$. To illustrate functors, we can define a functor $F: {\cal C}_{R} \rightarrow {\cal C_T}$  from the Rubiks cube category, where each object is  Rubiks cube state and arrows are moves, to a topological category ${\cal C}_T$ where each object is an element of a topological space and arrows are paths that connect one point to another. Just as in the Rubiks cube, a topological space defines a groupoid under the connectivity relationship \citep{may1999concise}. 

\subsection{MDP or PSR homomorphisms define categories}

Let us consider some further examples of categories for RL, to help build more intuition. 
A Markov decision process \citep{DBLP:books/lib/Bertsekas05,WHITE19891,PUTERMAN1990331}is defined by a tuple $\langle S, A, \Psi, P, R \rangle$, where $S$ is a discrete set of states, $A$ is the discrete set of actions, $\Psi \subset S \times A$ is the set of admissible state-action pairs, $P: \Psi \times S \rightarrow [0,1]$ is the transition probability function specifying the one-step dynamics of the model, where $P(s,a,s')$ is the transition probability of moving from state $s$ to state $s'$ in one step under action $a$, and $R: \Psi \rightarrow \mathbb{R}$ is the expected reward function, where $R(s,a)$ is the expected reward for executing action $a$ in state $s$. 

Since Rubiks cube and Backgammon are highly simplified problems, not indicative of real-world problems like medical diagnosis, where internal body states are not directly perceivable, there has been much interest in partially observable MDPs. A predictive state representation (PSR), an example of a POMDP-type representation,  is (in the simplest case) a discrete controlled dynamical system, characterized by a finite set of actions $A$, and observations $O$. At each clock tick $t$, the agent takes an action $a_t$ and receives an observation $o_t \in O$. A {\em history} is defined as a sequence of actions and observations $h = a_1 o_1 \ldots a_k o_k$. A {\em test} is a possible sequence of future actions and observations $t = a_1 o_1 \ldots a_n o_n$. A test is successful if the observations $o_1 \ldots o_n$ are observed in that order, upon execution of actions $a_1 \ldots a_n$. The probability $P(t | h)$ is a prediction of that a test $t$ will succeed from history $h$.

We can easily define categories, whose objects are MDPs or PSRs, by defining the arrows of these categories to be the respective homomorphisms.  We can define a category MDP 
 homomorphisms \citep{DBLP:conf/ijcai/RavindranB03,DBLP:conf/nips/Rezaei-Shoshtari22} or predictive state representation (PSR) homormorphisms \citep{DBLP:conf/aaai/SoniS07}. These are actually categories in disguise, although they have so far not been viewed as such. Let us walk through the main ideas in category theory using the MDP or PSR homomorphism example.

 MDP homomorphisms can be viewed as a principled way of abstracting the state (action) set of an MDP into a ``simpler" MDP that nonetheless preserves some important properties, usually referred to as the stochastic substitution property (SSP). This is a special instance of the general bisimulation property that holds among coalgebras, which we defined above. 

\begin{definition} 
\label{mdp-hom}
An MDP homomorphism \cite{DBLP:conf/ijcai/RavindranB03} from object  $M = \langle S, A, \Psi, P, R \rangle$ to $M' = \langle S', A', \Psi', P', R' \rangle$, denoted $h: M \twoheadrightarrow M'$, is defined by a tuple of surjections $\langle f, \{g_s | s \in S \} \rangle$, where $f: S \twoheadrightarrow S', g_s: A_s \twoheadrightarrow A'_{f(s)}$, where $h((s,a)) = \langle f(s), g_s(a) \rangle$, for $s \in S$, such that the stochastic substitution property and reward respecting properties below are respected: 
\begin{eqnarray} 
P'(f(s), g_s(a), f(s')) = \sum_{s" \in [s']_f} P(s, a, s") \\
R'(f(s), g_s(a)) = R(s, a) 
\end{eqnarray} 
\end{definition} 

We can construct a commutative diagram to illustrate MDP homomorphisms. Such diagrams will be used extensively in Section~\ref{topos-url} to illustrate properties of the topos of action-value functions. Intuitively, we can ``simulate" the base MDP in the ``abstract" MDP by either taking the abstraction step through the left $f$ down arrow, and then ``running" the abstract MDP for one step, or equivalently, running the base MDP for one step, and then abstracting the resultant state $t$. One can define an equivalent commutative diagram for the reward-respecting abstractions, which should satisfy the equation $R(.,a) = R'(., a) \circ f$. We can define such commutative diagrams in RL in terms of more advanced categorical constructions, such as (co)limits and (co)equalizers, which we will discuss in more detail in Section~\ref{topos-url}. Briefly, these constructions are formally {\em diagram} functors $F: {\cal J} \rightarrow {\cal C}$ from some finite category ${\cal J}$ into the category of choice ${\cal C}$. One can show that, for instance, the category ${\cal C_Q}$ of action-value $Q$-functions is a {\em topos} \citep{maclane1992sheaves}, which means it has all (co)limits. These properties will play a significant role in URL. 

\begin{center}
 \label{gdcarrow}
%https://q.uiver.app/#q=WzAsNCxbMCwwLCJVIl0sWzAsMiwiViJdLFsyLDAsIlUnIl0sWzIsMiwiViciXSxbMCwxLCJmIl0sWzAsMiwiaCIsMl0sWzIsMywiZiciLDJdLFsxLDMsImciXV0=
\begin{tikzcd}
	s && {t} \\
	\\
	s' && {t'}
	\arrow["{P(.,a)}"', from=1-1, to=1-3]
	\arrow["f", from=1-1, to=3-1]
	\arrow["{f'}"', from=1-3, to=3-3]
	\arrow["{P'(.,a)}", from=3-1, to=3-3]
\end{tikzcd}
 \end{center}

Given this definition, the following result is straightforward to prove. 

\begin{theorem}
The category ${\cal C}_{\mbox{MDP}}$ is defined as one where each object ${\cal M}$ is defined by an MDP, and the arrows are given by MDP homomorphisms are as defined in Definition~\ref{mdp-hom}. 
\end{theorem}

{\bf Proof:} Note that the composition of two MDP homomorphisms $h: M_1 \rightarrow M_2$ and $h': M_2 \rightarrow M_3$ is once again an MDP homomorphism $h' \ h: M_1 \rightarrow M_3$. The identity homomorphism is easy to define, and MDP homomorphisms, being surjective mappings, obey associative properties. $\qed$

Similarly, we now define the category ${\cal C}_{\mbox{PSR}}$ of predictive state representations \citep{singh-uai04}, based on the notion of homomorphism defined for PSRs proposed in  \citep{DBLP:conf/aaai/SoniS07}.   Note how the PSR straightforwardly generalizes the notion of a diversity-based representation that we explored above for the deterministic finite state environments to the stochastic setting. 

A state $\psi$ in a PSR is a vector of predictions of a suite of {\em core tests} $\{q_1, \ldots, q_k \}$. The prediction vector $\psi_h = \langle P(q_1 | h) \ldots P(q_k | h) \rangle$ is a sufficient statistic, in that it can be used to make predictions for any test. More precisely, for every test $t$, there is a $1 \times k$ projection vector $m_t$ such that $P(t | h) = \psi_h . m_t$ for all histories $h$. The entire predictive state of a PSR can be denoted $\Psi$. 

\begin{definition} 
\label{psr-homo}
The category ${\cal C}_{\mbox{PSR}}$ defined by PSR objects, the morphism from object $\Psi$ to another  $\Psi'$ is defined by a tuple of surjections $\langle f, v_\psi(a)\rangle$, where $f: \Psi \rightarrow \Psi'$ and $v_\psi: A \rightarrow A'$ for all prediction vectors $\psi \in \Psi$ such that 
\begin{equation}
    P(\psi' | f(\psi), v_\psi(a)) = P(f^{-1}(\psi') | \psi, a) 
\end{equation}
for all $\psi' \in \Psi, \psi \in \Psi, a \in A$. 
\end{definition} 

\begin{theorem}
The category ${\cal C}_{\mbox{PSR}}$ is defined by making each object $c$ represent a PSR, where the morphisms between two PSRs $h: c \rightarrow d$ is defined by the PSR homomorphism defined in \cite{DBLP:conf/aaai/SoniS07}. 
\end{theorem}

 {\bf Proof:} Once again, given the homomorphism definition in Definition~\ref{psr-homo}, the UDM category ${\cal P}_{\mbox{PSR}}$ is easy to define, given the surjectivity of the associated mappings $f$ and $v_\psi$. $\qed$

 For other examples of categories for decision making, we refer the reader to our previous paper on universal decision models \citep{sm:udm}. 

 \subsection{Category ${\cal C_Q}$ of Action-Value Functions} 

 One particularly simple example is worth highlighting as we will use it below to explain the concept of a topos, and to illustrate universal constructions in URL. 

 \begin{definition}  \label{transformercat}
    The category ${\cal C_Q}$ of action-value functions is defined as the following category: 
    \begin{itemize}
        \item The objects $\mbox{Obj}({\cal C_Q})$ are defined as Q-functions $f,g$, each mapping the state action set $S \times A,S' \times A'$ in an MDP (or PSR etc. into real numbers $f(s,a), f'(s',a') \in \mathbb{R}$, where $s \in S, s' \in S'$. 

        \item The arrows of the category ${\cal C_Q}$ are defined as commutative diagrams of the type shown above (which compose horizontally by adding boxes). 
 
 \begin{center}
 \label{gdcarrow2}
%https://q.uiver.app/#q=WzAsNCxbMCwwLCJVIl0sWzAsMiwiViJdLFsyLDAsIlUnIl0sWzIsMiwiViciXSxbMCwxLCJmIl0sWzAsMiwiaCIsMl0sWzIsMywiZiciLDJdLFsxLDMsImciXV0=
\begin{tikzcd}
	S \times A && {S' \times A'} \\
	\\
	\mathbb{R} && {\mathbb{R}}
	\arrow["h"', from=1-1, to=1-3]
	\arrow["f", from=1-1, to=3-1]
	\arrow["{f'}"', from=1-3, to=3-3]
	\arrow["g", from=3-1, to=3-3]
\end{tikzcd}
 \end{center}
 In the above diagram, the value functions $f,f'$ correspond to some MDPs,  and $g,h$ are mappings between their corresponding states or values that make the diagram commute.
    \end{itemize}
\end{definition} 
\subsection{Functor Categories and Yoneda Embeddings}
We now define a novel characterization of isomorphism in MDPs or PSRs using functor categories, in particular, Yoneda embeddings that arise from the application of the Yoneda Lemma \citep{maclane:71}. Given two functors $F, G: {\cal C} \rightarrow {\cal D}$, the {\em natural transformation} $\eta : F \Rightarrow G$ consists of a collection of arrows $\alpha_c: Fc \rightarrow Gc$ in category ${\cal D}$ such that the commutative diagram below composes (meaning $G f \circ \alpha_c = \alpha_{c'} \circ Ff$):
\[\begin{tikzcd}
	Fc &&& Gc \\
	\\
	{Fc'} &&& {Gc'}
	\arrow["{\alpha_c}", from=1-1, to=1-4]
	\arrow["Ff"', from=1-1, to=3-1]
	\arrow["{\alpha_{c'}}"', from=3-1, to=3-4]
	\arrow["Gf", from=1-4, to=3-4]
\end{tikzcd}\]
We can interpret each arrow $Ff$ or $Gf$ in terms of an MDP or PSR homomorphism, for the category ${\cal C_V}$. This interpretation will take on additional meaning once we introduce the universal constructions, such as (co)limits. 

\begin{definition} 
    A functor category ${\cal C}^{\cal D}$ is defined as the category where every object is a functor $F: {\cal D} \rightarrow {\cal C}$, and whose arrows are natural transformations between two functors $F$ and $G$ as illustrated above. 
\end{definition} 

One particular functor category of great significance for URL is the category of covariant (contravariant) set-valued functors ${{\bf Set}}^{\cal C}$ (${{\bf Set}}^{C^{op}}$), respectively,  where each object is a covariant (contravariant) $F: {\cal C} \rightarrow {\bf Set}$ ($F: {\cal C}^{op} \rightarrow {\bf Set}$). This functor category plays a key role in the celebrated Yoneda Lemma \citep{maclane:71,yoneda-end}, which shows that objects in a category can be defined completely (up to isomorphism) purely  from their interactions with other objects. We can use the Yoneda Lemma to give a new definition of isomorphism in MDPs and PSRs, and other dynamical system models in URL. 

In particular, the categories  ${\cal C}_{MDP}$ and ${\cal C}_{PSR}$ can be ``faithfully" embedded in the category of sets, denoted ${\bf Sets}$, using the Yoneda embedding $c \rightarrow C(c, .): {\cal C} \rightarrow {\bf Sets}^{\cal C}$ (referred to as the {\em covariant} embedding): 

\begin{lemma}
{\bf {Yoneda lemma}}: For any functor $F: C \rightarrow {\bf Set}$, whose domain category $C$ is ``locally small" (meaning that the collection of morphisms between each pair of objects forms a set), any object $c$ in $C$, there is a bijection $\mbox{Hom}(C(c, -), F) \simeq Fc$  that defines a natural transformation $\alpha: C(c, -) \Rightarrow F$ to the element $\alpha_c(1_c) \in Fc$. This correspondence is natural in both $c$ and $F$. 
\end{lemma}

To give the complete characterization of MDPs and PSRs in terms of Yoneda embeddings, we can use the following definition: 

\begin{definition} 
    The covariant Yoneda embedding of an MDP is defined as the set-valued functor $F_{MDP}: {\cal C}_{MDP} \rightarrow {\bf Set}$, where any given MDP ${\cal M}$ is mapped into the set of MDP homomorphisms going out of it, namely ${\cal M} \rightarrow {\cal C}_{MDP}({\cal M}, -)$.  The contravariant Yoneda embedding of an MDP is defined as the set-valued functor $F^{op}_{MDP}: {\cal C}^{op}_{MDP} \rightarrow {\bf Set}$, where any given MDP ${\cal M}$ is mapped into the set of MDP homomorphisms going into it, namely ${\cal M} \rightarrow {\cal C}_{MDP}(-,{\cal M})$.
\end{definition} 

Similarly, we can define contravariant and covariant Yoneda embeddings of PSRs as well. 

\begin{definition} 
    The covariant Yoneda embedding of a PSR is defined as the set-valued functor $F_{MDP}: {\cal C}_{PSR} \rightarrow {\bf Set}$, where any given PSR $\Psi$ is mapped into the set of PSR homomorphisms going out of it, namely $\Psi \rightarrow {\cal C}_{PSR}(\Psi, -)$.  The contravariant Yoneda embedding of a PSR is defined as the set-valued functor $F^{op}_{PSR}: {\cal C}^{op}_{PSR} \rightarrow {\bf Set}$, where any given PSR $\Psi$ is mapped into the set of PSR homomorphisms going into it, namely $\Psi \rightarrow {\cal C}_{MDP}(-,\Psi)$.
\end{definition} 

\section{Exact Solution Methods for MDPs as Functor Categories}
\label{dpfuncat}
We now show that exact solution methods for solving MDPs, such as value iteration and policy iteration, as well as simulation-based methods, such as Monte-Carlo methods and TD-learning, can be viewed as functors mapping from the category ${\cal C}_{MDP}$ to the category ${\cal C}_V$ of value functions. 

\subsection{Value Iteration as a Coalgebra} 

One way to solve an MDP is to construct the optimal value function $V^*$, or action value function $Q^*(s,a)$.  Since these functions are the fixed points of operators, which are final coalgebras, a natural strategy is to find the fixed point by successive  approximation.   The value iteration algorithm computes the next approximation $V^{t+1}$  by iteratively ``backing up" the current approximation: 

\begin{equation}
\label{value-iteration}
V^{t+1}(s) = T^*(V^t)(s) =  \max_{a} \left( R_{sa} + \gamma \sum_{a} P_{ss'}^a V^{t}(s') \right)
\end{equation}

Note how $T^*(V) = V$ is essentially a coalgebraic equation, specified by a functor $T^*$ that gives the $T^*$ dynamics of this coalgebra. The value iteration algorithm can be terminated when the difference between successive approximations is less than some tolerance parameter, that is  $|| V^{t+1} - V^t || \leq \epsilon$, where $||.||$ denotes the ``norm" or ``length" of a function.  It can be shown that the operator $T^*$ is a {\em contraction} in a Hilbert space, and hence the algorithm will asymptotically  converge \cite{bertsekas:rlbook}.\footnote{Intuitively, this property implies that at each iteration, the ``distance" between $V^t$ and $V^*$ shrinks, and hence asymptotically, the algorithm must converge.}   A similar procedure can be used to compute the optimal action value function $Q^*(s,a)$. A drawback of value iteration is that its convergence can be slow for  $\gamma$ near  $1$. Also, at each step, it is necessary to ``backup" values over the entire state space. This problem has been addressed by a number of approaches:  algorithms such as {\em real-time dynamic programming} (RTDP) and  ``simulation-based" reinforcement learning methods back up values over only a set of sample transitions \cite{bertsekas:rlbook}. 

\subsection{Policy Iteration as an Adjoint Functor} 
\label{pol-it}

\begin{algorithm}
\caption{The Policy Iteration Algorithm is a Functor $F_{PI}: {\cal C}_M \rightarrow {\cal C}_V$.} 

Set $\pi$ to some random initial policy. 

\begin{algorithmic}[1]
\REPEAT 

\STATE \label{policy-eval} Solve the ``Bellman" linear system of equations to determine $V^\pi$

\[ (I - \gamma P^\pi) V^\pi  = R^\pi  \] 

\STATE Find the ``greedy" policy $\pi'$ associated with $V^\pi$: 

\[ \pi'(s) \in \mbox{argmax}_a \left( \sum_{s'} P_{ss'}^a \left(R_{ss'}^a + \gamma V(s') \right) \right) \] 

\IF{$\pi'  \neq  \pi$}

\STATE Set $\pi \leftarrow \pi'$ and return to Step~\ref{policy-eval}. 

\ELSE 

\STATE Set {\bf done} $\leftarrow$ {\bf true}. 

\ENDIF 

\UNTIL{{\bf done}}. 

Return $\pi$ as the optimal discounted policy for MDP $M$. 

\end{algorithmic}
\end{algorithm} 

In terms of the Figure~\ref{fig:mdpfunctor}, we can view policy iteration as a {\em adjoint} pair of functions, one that maps from a policy $\pi$ to its associated value function $V^\pi$, and then another functor that maps from a value function $V^\pi$ back to the greedy policy $\pi'$ associated with $V^\pi$. 

The {\em policy iteration} algorithm was introduced by Howard \cite{howard:dp}. In this procedure, at each iteration, the decision maker ``evaluates" a specific policy $\pi$, finding its associated value function $V^\pi$. Then, the associated ``greedy" policy $\pi'$ associated with $V^\pi$ is computed, which deviates from $\pi$ for one step by finding the action that maximizes the one-step reward and then follows $\pi$ subsequently. Howard proved that if $\pi$ is not the optimal policy, then the greedy policy associated with $V^\pi$ must improve on $\pi$.  As described here, the policy iteration algorithm assumes that value functions (and other functions such as rewards, transition models, and policies) can be stored exactly, using a ``table lookup" representation. 

\subsection{Linear Programming for MDPs as a Functor}

A third approach to exactly solving MDPs is based on {\em linear programming} \cite{NAZARETH198613}. The variables for the linear program are the values $V(i)$. The exact formulation is given below: 

\begin{definition} 
The linear program required to solve an MDP M is given as
\begin{eqnarray*}
\mbox{Variables} &:& V(1), \ldots, V(n) \\ 
\mbox{Minimize } &:& \sum_s \alpha_s V(s) \\
\mbox{Subject to}  &:& V(s) \geq  \sum_{s'} P_{ss'}^a \left( R_{ss'}^a + \gamma V(s') \right) , \ \ \forall s \in S,  a \in A, 
\end{eqnarray*}
where $\alpha$ is a state relevance weight vector whose weights are all positive. 
\end{definition} 

There is one constraint for each state $s$ and action $a$, thus leading to an intractable set of constraints in problems with an exponential state space. 

\section{Simulation-based and Reinforcement Learning Methods as Functors} 
\label{rlfuncat}

We now turn to briefly describe a class of approximation methods for solving MDPs, which retain the restriction of representing functions exactly, but require only sample transitions $(s_t, a_t, r_t, s'_t)$ instead of true knowledge of the MDP. Such methods can be generically referred to as {\em simulation-based} methods, and are the topic of much study in various areas, including {\em approximate dynamic programming} \cite{bertsekas:rlbook} and {\em reinforcement learning} \cite{DBLP:books/lib/SuttonB98}. Two classes of methods will be described: Monte-Carlo methods, which approximate the exact return $V^\pi$ by summing the actual returns, and {\em temporal-difference learning} methods, which can be viewed as a biased version of Monte-Carlo methods. 

\subsection{Monte-Carlo Methods as Functors} 

Monte-Carlo methods have long been studied in a variety of fields, and there is a well-developed statistical theory underlying them. Monte-Carlo methods for solving MDPs are based on the simple idea that the value of a particular state $V^\pi(s)$ associated with a particular policy $\pi$ can be empirically determined by ``simulating" the policy $\pi$ on a given MDP $M$, and averaging the sum of rewards received. Monte-Carlo methods are simple to implement. As a statistical procedure, they have the attractive property of being {\em unbiased} estimators of the true value. Unfortunately, their variance can be high. A more detailed discussion of Monte-Carlo methods for MDPs is  given in \cite{bertsekas:rlbook,DBLP:books/lib/SuttonB98}. Note that we can represent Monte-Carlo evaluation of a policy $\pi$ as a functor from the category ${\cal C}_{MDP}$ of MDPs into the category ${\cal C}_V$ of value functions. 

\begin{algorithm}
\caption{Monte Carlo evaluation as a functor $F_M: {\cal C}_M \rightarrow {\cal C}_V$.} 

\begin{algorithmic} 
\STATE Given an MDP ${\cal M}$ and a policy $\pi$ for drawing samples from ${\cal M}$.
\FOR{$i=1$ to $N$} 

\STATE Set the step counter $t=0$, initial state $s_t  = s$, and $\hat{V}_i = 0$. 

\STATE Set the action $a = \pi(s_t)$, and ``execute" action $a$. 

\STATE Let the sample reward received be $r_t$.  Set $t \leftarrow t+1$. 

\STATE Set $\hat{V}_i = \hat{V}_i + r_t$. 

\IF{terminated}

\STATE Set $\hat{V}_i = \frac{1}{t} \hat{V}_i$. 

\ENDIF 

\ENDFOR 

Return $\hat{V}^\pi(s) = \frac{1}{N} \sum_{i=1}^N \hat{V}_i$. 
\end{algorithmic} 
\end{algorithm} 

\subsection{Temporal-Difference Learning as a Functor} 

Temporal-difference learning (TD) methods \cite{DBLP:books/lib/SuttonB98} exploit the property that the value of a state $V^\pi(s)$ can be estimated as the sum of the immediate reward received, $r_t$, and a {\em biased} estimate of the value at the next state. If we think of TD-learning as a sample-based variant of value iteration, then it follows straightforwardly that it can be modeled as a coalgebra or endofunctor on the category of value functions ${\cal C}_V$. In other words, given some guess of a current value function $\hat{V}$, which is an object in ${\cal C}_V$, an update step of TD simply returns a revised value function $\hat{V}'$. The simplest TD($0$) algorithm can be written as: 

\begin{equation}
\label{td-policy-eval} 
\hat{V}^\pi_{t+1}(s) \leftarrow (1 - \alpha_t) \hat{V}^\pi_t(s) + \alpha_t \left(r_t  + \hat{V}^\pi_t(s) \right) 
\end{equation} 

Here, $\alpha_t \in (0,1)$ is a time-varying ``learning rate" that  averages the estimate over multiple samples. As in the Monte-Carlo procedure, the TD($0$) algorithm also only requires a simulation of an MDP, and asynchronously evaluates the value of states in an MDP over sampled trajectories.    Readers familiar with the theory of stochastic approximation \cite{kushner2003stochastic}  may recognize the TD update rule given above as similar to the  {\em Robbins-Munro} method of finding the roots of a function. Indeed, a rigorous convergence analysis of TD learning has been made drawing on the theory of stochastic approximation  \cite{borkar}.  A TD-based method for learning action values called Q-learning was introduced by Waktins, whose convergence was analyzed by \citet{tsitsiklis}.  Q-learning estimates the optimal action value function using the following learning rule: 

\begin{equation}
\label{q-learning}
Q_t(s,a) \leftarrow (1 - \alpha_t) Q_t(s,a) + \alpha_t (r_t + \gamma \max_{a'} Q_t(s',a') ) 
\end{equation} 

Q-learning is sometimes referred to as an ``off-policy" learning algorithm since it estimates the optimal action value function $Q^*(s,a)$, while simulating the MDP using any policy, such as a random walk. In practice, a greedy policy is used that picks  actions at each state with the highest $Q(x,a)$,  occasionally choosing random actions to ensure sufficient exploration. 

\section{Deep RL as  Diagrams over Coalgebras}
\label{deeprl}

Getting back to our earlier remark that in URL, architectures are specified as functor diagrams $F: {\cal J} \rightarrow {\cal C}$, we can concretize this view with the example of deep RL.  We now show how we can model deep RL as a functor diagram, which can be seen as a variant of the general functorial characterization of backpropagation \citep{DBLP:conf/lics/FongST19}.  Our principal goal in this section to extend the categorical framework for deep learning proposed in \cite{DBLP:conf/lics/FongST19} to model deep RL.   We first introduce a brief review of symmetric monoidal categories, where objects and arrows can be ``multiplied" as a tensor product. Diagrammatically, these refer to running processes in parallel to compute outputs. This interpretation will be of value when we discuss asynchronous computation in URL. 

\subsection{Symmetric Monoidal Categories}

\label{smc-cat}

\begin{figure}[h]
    \centering
    \includegraphics[width=0.75\linewidth]{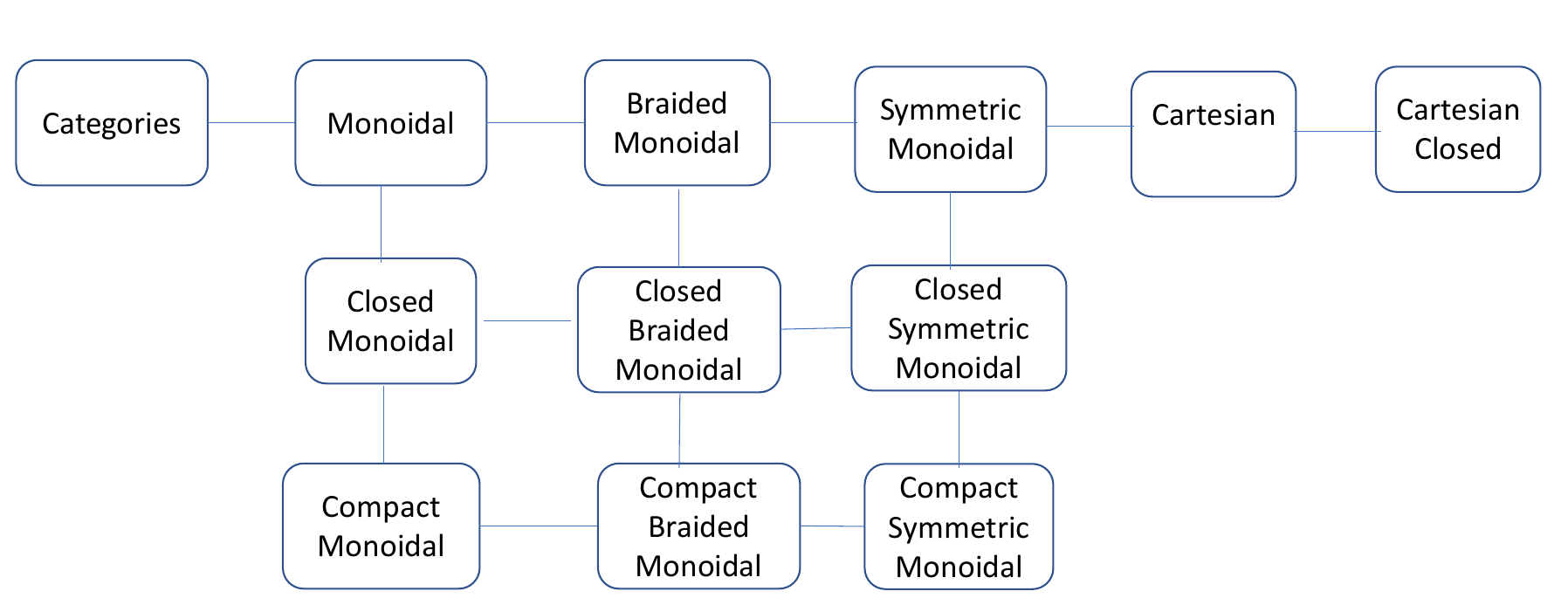}
    \caption{Structure of monoidal categories.}
    \label{monoidal-cats}
\end{figure}

 We assume the reader understands the basics of symmetric monoidal categories, which we briefly review below (see Figure~\ref{monoidal-cats}). Good introductions are available in a number of textbooks \cite{maclane:71,richter2020categories}. A brief introduction to some basic category theory suitable for causal inference is in my previous paper \cite{DBLP:journals/entropy/Mahadevan23}.  Detailed overviews of symmetric monoidal categories appear in many books, and our definitions are based on \cite{richter2020categories}. 

\begin{definition} 
    A {\bf monoidal category} is a category {\cal C} together with a functor $\otimes: {\cal C} \times {\cal C} \rightarrow {\cal C}$, an identity object $e$ of {\cal C} and natural isomorphisms $\alpha, \lambda, \rho$ defined as: 

    \begin{eqnarray*}
        \alpha_{C_1, C_2, C_3}: C_1 \otimes (C_2 \otimes C_3) & \cong & (C_1 \otimes C_2) \otimes C_2, \ \ \mbox{for all objects} \ \ C_1, C_2, C_3 \\
        \lambda_C: e \otimes C & \cong & C, \ \ \mbox{for all objects} \  \ C  \\
        \rho: C \otimes e & \cong & C, \ \  \mbox{for all objects} \ C
    \end{eqnarray*}
\end{definition} 

The natural isomorphisms must satisfy coherence conditions called the ``pentagon" and ``triangle" diagrams \cite{maclane:71}. An important result shown in \cite{maclane:71} is that these coherence conditions guarantee that all well-formed diagrams must commute. 

There are many natural examples of monoidal categories, the simplest one being the category of finite sets, termed {\bf FinSet} in \cite{Fritz_2020}, where each object $C$ is a set, and the tensor product $\otimes$ is the Cartesian product of sets, with functions acting as arrows. Deterministic causal models can be formulated in the category {\bf FinSet}. Other examples include the category of sets with relations as morphisms, and the category of Hilbert spaces \cite{Heunen2019}.  The category {\bf FinSet}  has other properties, principally that the $\otimes$ is actually a product (in that it satisfies the universal property of products in categories, and is formally a limit). Not all monoidal categories satisfy this property. Sets are also Cartesian closed categories, meaning that there is a right adjoint to the tensor product, which represents exponential objects, and is often referred to as the ``internal hom" object. Markov categories are monoidal categories, where the identity element $e$ is also a terminal object, meaning there is a unique ``delete" morphism $d_e: X \rightarrow e$ associated with each object $X$. This property can be used to show that projections of tensor products $X \otimes Y$ exist, but they do not satisfy the universal property. Markov categories do not satisfy uniform copying. 

\begin{definition} 
    A {\bf symmetric monoidal category} is a monoidal category $({\cal C}, \otimes, e, \alpha, \lambda, \rho)$ together with a natural isomorphism 

    \begin{eqnarray*}
       \tau_{C_1, C_2}: C_1 \otimes C_2 \cong C_2 \otimes C_1, \ \ \mbox{for all objects} \ \ C_1, C_2
    \end{eqnarray*}
    where $\tau$ satisfies the additional conditions: for all objects $C_1, C_2$ $\tau_{C_2, C_1} \circ \tau_{C_1, C_2} \cong 1_{C_1 \otimes C_2}$, and for all objects $C$, $\rho_C = \lambda_C \circ \tau_{C, e}: C \otimes e \cong C$. 
\end{definition} 

An additional hexagon axiom is required to ensure that the $\tau$ natural isomorphism is compatible with $\alpha$.  The $\tau$ operator is called a ``swap" in Markov categories \cite{Fritz_2020}. 

\subsection{Category of Compositional Reinforcement Learners }

\cite{DBLP:conf/lics/FongST19} give an elegant characterization of the well-known backpropagation algorithm that serves as the ``workhorse" of deep learning as a functor  over symmetric monoidal categories. In such categories, objects can be ``multiplied": for example, sets form a symmetric monoidal category as the Cartesian product of two sets defines a multiplication operator. A detailed set of coherence axioms are defined for monoidal categories (see \cite{maclane:71} for details), which we will not go through, but they ensure that multiplication is associative, as well as that there are identity operators such that $I \otimes A \simeq A$ for all objects $A$, where $I$ is the identity object. 

\begin{figure}
    \centering
    \includegraphics[scale=0.4]{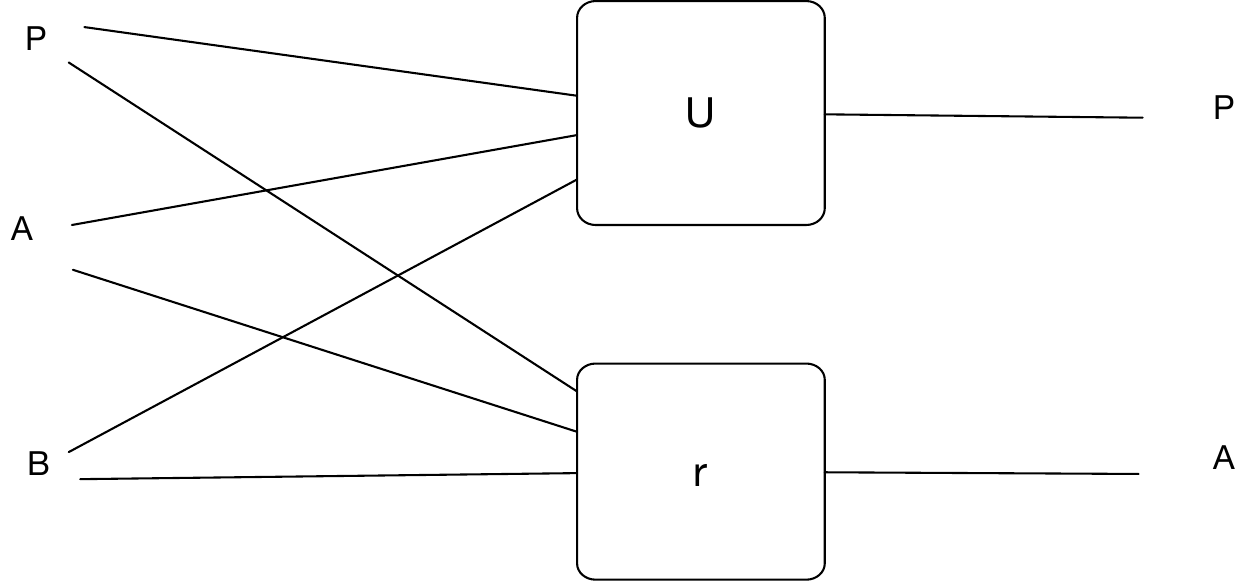}
    \caption{A Reinforcement Learner in the symmetric monoidal category {\tt Learn} is defined as a morphism.}
    \label{learncat}
\end{figure}

\begin{definition} \cite{DBLP:conf/lics/FongST19}
    The symmetric monoidal category {\bf Learn} is defined as a collection of objects that define sets, and a collection of an equivalence class of learners. Each learner is defined by the following $4$-tuple (see Figure~\ref{learncat}). 
    
\begin{itemize}
    \item A parameter space $P$ that defines the representable policies. 

    \item An implementation function $I: P \times A \rightarrow B$ that computes the (action)value function. 

    \item An update function $U: P \times A \times B \rightarrow P$ that updates the policy parameters (or action-value parameters). 

    \item A request function $r: P \times A \times B \rightarrow A$ (which can be seen as the TD backup step). 
\end{itemize}

Note that it is the request function that allows learners to be composed, as each request function transmits information back upstream to earlier learners what output they could have produced that would be more ``desirable". This is exactly analogous to the TD backup procedure studied in \citep{bertsekas:rlbook,DBLP:books/lib/SuttonB98}. This algebraic characterization of the backpropagation algorithm clarifies its essentially compositional nature 

Two reinforcement learners  $(P, I, U, R)$ and $(P', I', U', r')$ are equivalent if there is a bijection $f: P \rightarrow P'$ such that the following identities hold for each $p \in P, a \in A$ and $b \in B$. 

\begin{itemize}
    \item $I'(f(p), a) = I(p, a)$. 

    \item $U'(f(p), a, b) = f(U(p, a, b))$. 

    \item $r'(f(p), a, b) = r(p, a, b)$
\end{itemize}
    
\end{definition} 

Typically, in deep RL trained with neural networks, the parameter space $P = \mathbb{R}^N$ where the neural network has $N$ parameters. The implementation function $I$ represents the ``feedforward" component, and the request function represents the ``backpropagation" component. The update function represents the change in parameters as a result of processing a training example $(a, f(a)) \in A \times B$, which corresponds to the TD procedure described previously in Section~\ref{rlintro}. 

\begin{figure}[t]
    \centering
    \includegraphics[scale=0.4]{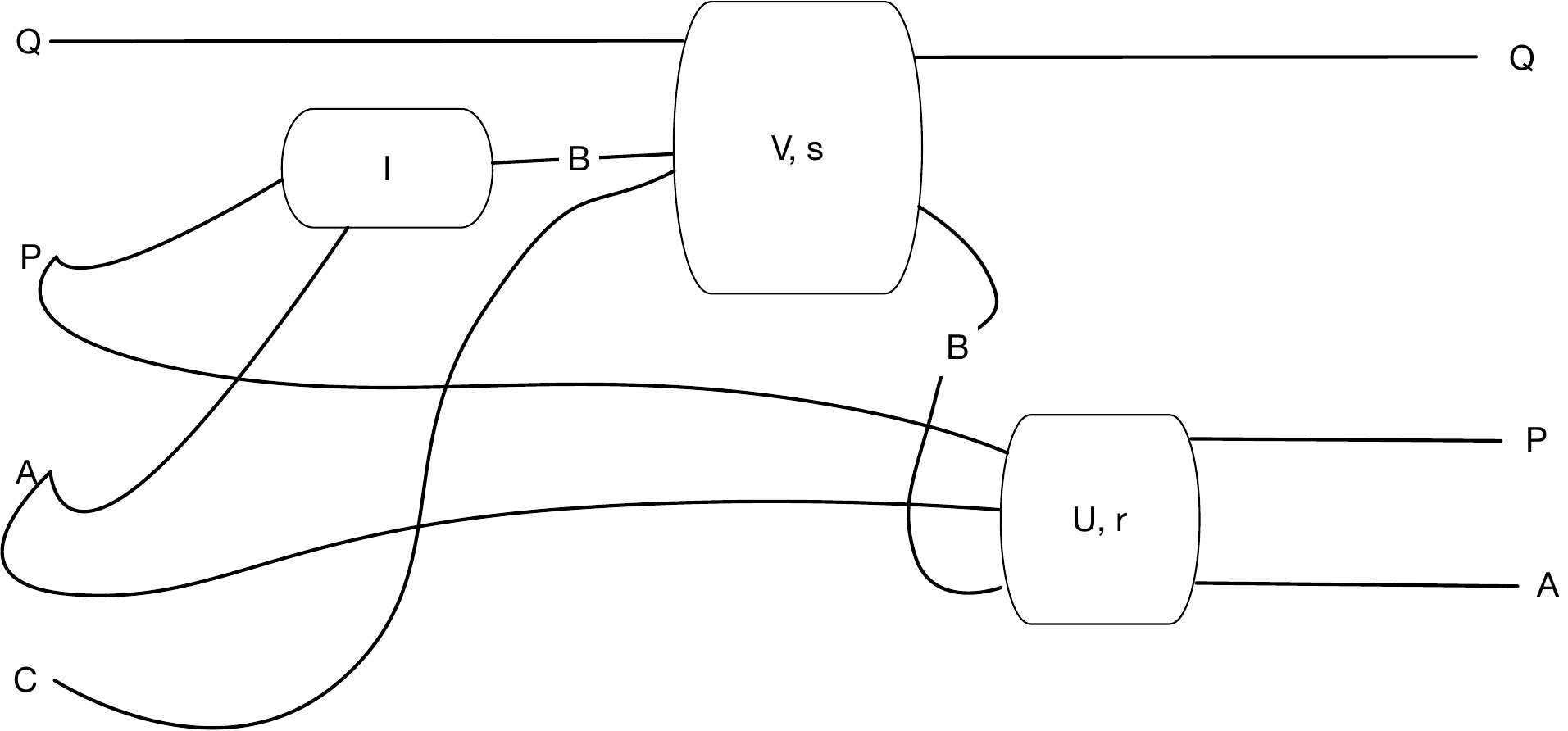}
    \includegraphics[scale=0.4]{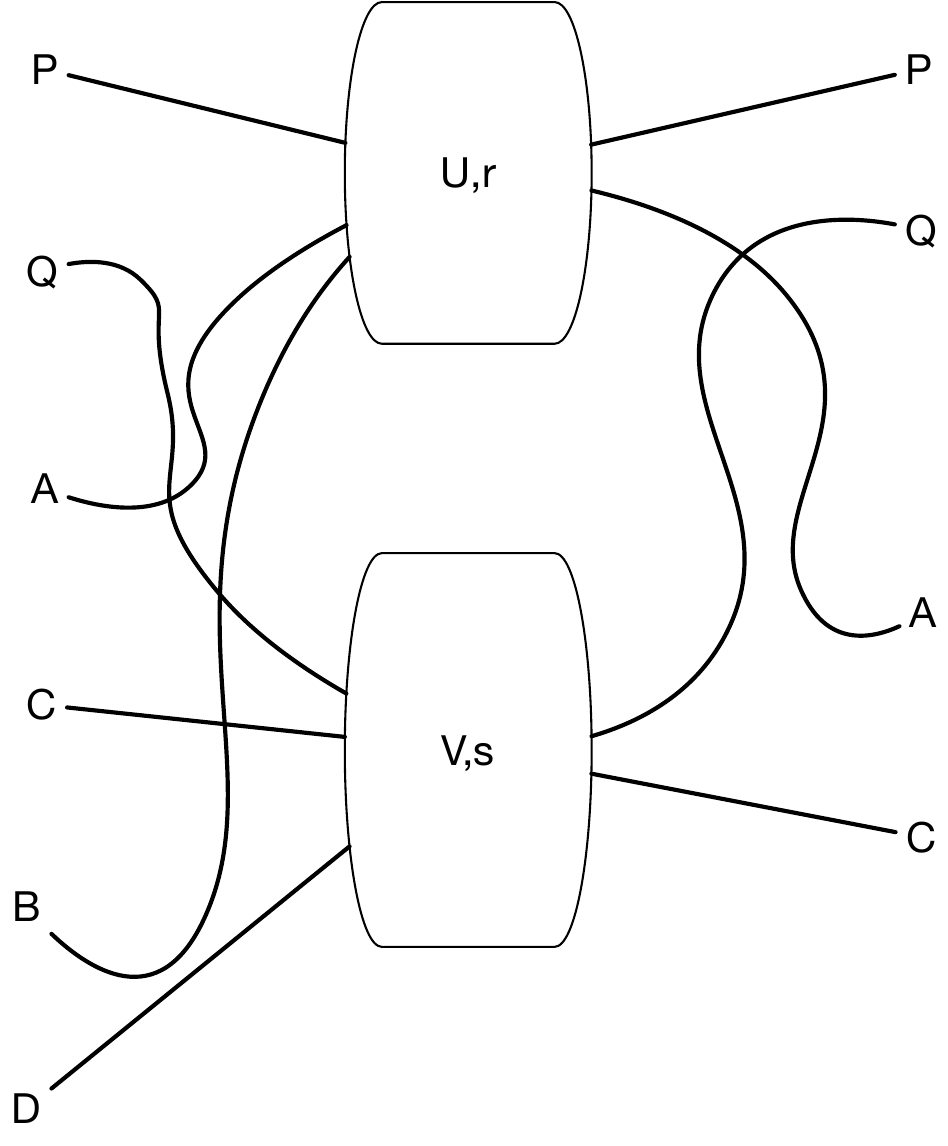}
    \caption{Sequential and parallel composition of two reinforcemnt learners in the symmetric monoidal category {\tt Learn}.}
    \label{seqlearn}
\end{figure}

Each reinforcement learner can be combined in sequentially and in parallel (see Figure~\ref{seqlearn}), both formally using the operations of composition $\circ$ and tensor product $\otimes$ in the symmetric monoidal category {\tt Learn}, and equivalently in terms of string diagrams. For clarity, let us write out the compositional rule for a pair of learners

\[ A \xrightarrow[]{(P, I, U, r)} B \xrightarrow[]{(Q, J, V, s)} C \]

The composite learner $A \rightarrow C$ is defined as $(P \times Q, I \cdot J, U \cdot V, r \cdot s)$, where the composite implementation function is

\[ (I \cdot J)(p, q, a) \coloneqq J(q, I(p, a)) \]

and the composite update function is 

\[ U \cdot V(p, q, a, c) \coloneqq \left( U(p, a, s(q, I(p, a), c) \right), V(q, I(p, a), c) \]

and the composite request function is 

\[ (r \cdot s)(p, q, a, c) \coloneqq r(p, a, s(q, I(p, a), c)). \]

\subsection{Backpropagation as a Functor}

We can define the backpropagation procedure as a functor that maps from the category {\tt Para} to the category {\tt Learn}. Functors can be viewed as a generalization of the notion of morphisms across algebraic structures, such as groups, vector spaces, and graphs. Functors do more than functions: they not only map objects to objects, but like graph homomorphisms, they need to also map each morphism in the domain category to a corresponding morphism in the co-domain category. Functors come in two varieties, as defined below. The Yoneda Lemma, in its most basic form, asserts that any set-valued functor $F: {\cal C} \rightarrow {\bf Sets}$ can be universally represented by a {\em representable functor} ${\cal C}(-, x): {\cal C}^{op} \rightarrow {\bf Sets}$. 

The {\em functoriality} axioms dictate how functors have to be behave: 

\begin{itemize} 

\item For any composable pair $f, g$ in category $C$, $Fg \cdot Ff = F(g \cdot f) $.

\item For each object $c$ in $C$, $F (1_c) = 1_{Fc}$.

\end{itemize} 

Note that the category {\tt Learn} is ambivalent as to what particular learning method is used. To define a particular learning method, such as backpropagation, we can define a category whose objects define the parameters of the particular learning method, and then another category for the learning method itself.  We can define a functor from the category {\tt NNet} to the category {\tt Learn} that factors through the category {\tt Param}. Later in the next section, we show how to generlize this construction to simplicial sets. 

% https://q.uiver.app/#q=WzAsMyxbNCwwLCJMZWFybiJdLFswLDAsIk5OZXQiXSxbMiwxLCJQYXJhbSJdLFsxLDIsIkYiXSxbMiwwLCJMX3tcXGVwc2lsb24sIGV9Il0sWzEsMF1d
\[\begin{tikzcd}
	NNet &&&& Learn \\
	&& Param
	\arrow["F", from=1-1, to=2-3]
	\arrow["{L_{\epsilon, e}}", from=2-3, to=1-5]
	\arrow[from=1-1, to=1-5]
\end{tikzcd}\]

\begin{definition}  \cite{DBLP:conf/lics/FongST19}
    The category {\tt Param} defines a strict symmetric monoidal category whose objects are Euclidean spaces, and whose morphisms $f: \mathbb{R}^n \rightarrow \mathbb{R}^m$ are equivalence classes of differential parameterized functions. In particular, $(P, I)$ defines a Euclidean space $P$ and $I: P \times A \rightarrow B$ defines a differentiable parameterized function $A \rightarrow B$. Two such pairs $(P, I), (P', I')$ are considered equivalent if there is a differentiable bijection $f: P \rightarrow P'$ such that for all $p \in P$, and $a \in A$, we have that $I'(f'(p),a) = I(p,a)$. The composition of $(P, I): \mathbb{R}^n \rightarrow \mathbb{R}^m$ and $(Q, J): \mathbb{R}^n \rightarrow \mathbb{R}^m$ is given as 

    \[ (P \times Q, I \cdot J) \ \ \ \mbox{where} \ \ \ (I \cdot J)(p, q, a) = J(q, I(p, a)) \]

    The monoidal product of objects $\mathbb{R}^n$ and $\mathbb{R}^m$ is the object $\mathbb{R}^{n+m}$, whereas the monoidal product of morphisms $(P, I): \mathbb{R}^m \rightarrow \mathbb{R}^m$ and $(Q, J): \mathbb{R}^l \rightarrow \mathbb{R}^k$ is given as $(P \times Q, I \parallel J)$, where 

    \[ (I \parallel J) (p, q, a, c)  = \left( I(p, a), J(q, c) \right) \]

    Symmetric monoidal categories can also be braided. In this case, the braiding $\mathbb{R}^m \parallel \mathbb{R}^m \rightarrow \mathbb{R}^m \parallel \mathbb{R}^n $ is given as $(\mathbb{R}^0, \sigma)$ where $\sigma(a, b) = (b, a)$. 
\end{definition} 

The backpropagation algorithm can itself be defined as a functor over symmetric monoidal categories

\[ L_{\epsilon, e}: {\tt Param} \rightarrow {\tt Learn}\]

where $\epsilon > 0$ is a real number defining the learning rate for backpropagation, and $e(x,y): \mathbb{R} \times \mathbb{R} \rightarrow \mathbb{R}$ is a differentiable error function such that $\frac{\partial e}{\partial x}(x_0, -)$ is invertible for each $x_0 \in \mathbb{R}$. This functor essentially defines an update procedure for each parameter in a compositional learner. In other words,  the functor $L_{\epsilon, e}$ defined by backpropagation sends each parameterized function $I: P \times A \rightarrow B$ to the learner $(P, I, U_I,r_I)$

\[ U_I(p, a, b) \coloneqq p - \epsilon \nabla_p E_I(p, a, b) \]

\[ r_I(p, a, b) \coloneqq f_a(\nabla_a E_I(p, a, b)) \]

where $E_I(p, a, b) \coloneqq \sum_j e(I_j(p, a), b_j)$ and $f_a$ is a component-wise application of the inverse to $\frac{\partial e}{\partial x}(a_i, -)$ for each $i$. Here, $e$ can be defined as the TD error in $Q$-learning. 

Note that we can easily define functors that define other ways of doing parameterized updates, such as a stochastic approximation method \cite{rm} that updates each parameter using only the (noisy) value of the function at the current value of the parameter, and uses a gradual decay of the learning parameters to ``simulate" the process of taking gradients. These sort of stochastic approximation updates are now called ``zeroth-order" optimization in the deep learning literature. 

 \subsection{Backpropagation as a Coalgebra}

 We describe deep RL now not just as a functor, but rather as an endofunctor that maps an object in a category {\tt Param} of parameters into a new object in the same category as a result of doing a gradient update. We can now formally define backpropagation as a coalgebra over the caetgory {\tt Param} as follows.  Recall that the category {\tt Param} defines a strict symmetric monoidal category whose objects are Euclidean spaces, and whose morphisms $f: \mathbb{R}^n \rightarrow \mathbb{R}^m$ are equivalence classes of differential parameterized functions. To see why the backpropagation algorithm can be defined as an endofunctor over the symmetric monoidal category {\tt Param}, recall from the previous section that backpropagation was viewed as a functor from the cateory {\tt Param} to the cateogry {\tt Learn}. 

\[ L_{\epsilon, e}: {\tt Param} \rightarrow {\tt Learn}\]

where $\epsilon > 0$ is a real number defining the learning rate for backpropagation, and $e(x,y): \mathbb{R} \times \mathbb{R} \rightarrow \mathbb{R}$ is a differentiable error function such that $\frac{\partial e}{\partial x}(x_0, -)$ is invertible for each $x_0 \in \mathbb{R}$. This functor essentially defines an update procedure for each parameter in a compositional learner. In other words,  the functor $L_{\epsilon, e}$ defined by backpropagation sends each parameterized function $I: P \times A \rightarrow B$ to the learner $(P, I, U_I,r_I)$

\[ U_I(p, a, b) \coloneqq p - \epsilon \nabla_p E_I(p, a, b) \]

\[ r_I(p, a, b) \coloneqq f_a(\nabla_a E_I(p, a, b)) \]

where $E_I(p, a, b) \coloneqq \sum_j e(I_j(p, a), b_j)$ and $f_a$ is a component-wise application of the inverse to $\frac{\partial e}{\partial x}(a_i, -)$ for each $i$. 

But a simpler and we argue more elegant characterization of backpropagation is to view it as a coalgebra or dynamical system defined by an endofunctor on {\tt Param}. Here, we view the inputs $A$ and outputs $B$ as the input ``symbols" and output produced by a dynamical system. The actual process of updating the parameters need not be defined as ``gradient descent", but it can involve any other functor (as we saw earlier, it could involve a stochastic approximation method \cite{borkar}). Our revised definition of backpropagation as an endofunctor follows. Note that this definition is generic, and applies to virtually any approach to building foundation models that updates each object to a new object in the category {\tt Param} as a result of processing a data instance. 

  \begin{definition}
{\bf Backpropagation} defines an $F_B$-coalgebra over the symmetric monoidal category {\tt Param}, specified by an endofunctor $X \rightarrow F_B(X)$ defined as 

\[ F_B(X) = A \times B \times X\]
  \end{definition}

Note that in this definition, the endofunctor $F_B$ takes an object $X$ of {\tt Param}, which is a set of network weights of a generative AI model, and produces a new set of weights, where $A$ is the ``input" symbol of the dynamical system and $B$ is the output symbol. 

\subsection{Zeroth-Order Deep Learning using Stochastic Approximation}

\begin{figure}
    \centering
    \includegraphics[scale=0.6]{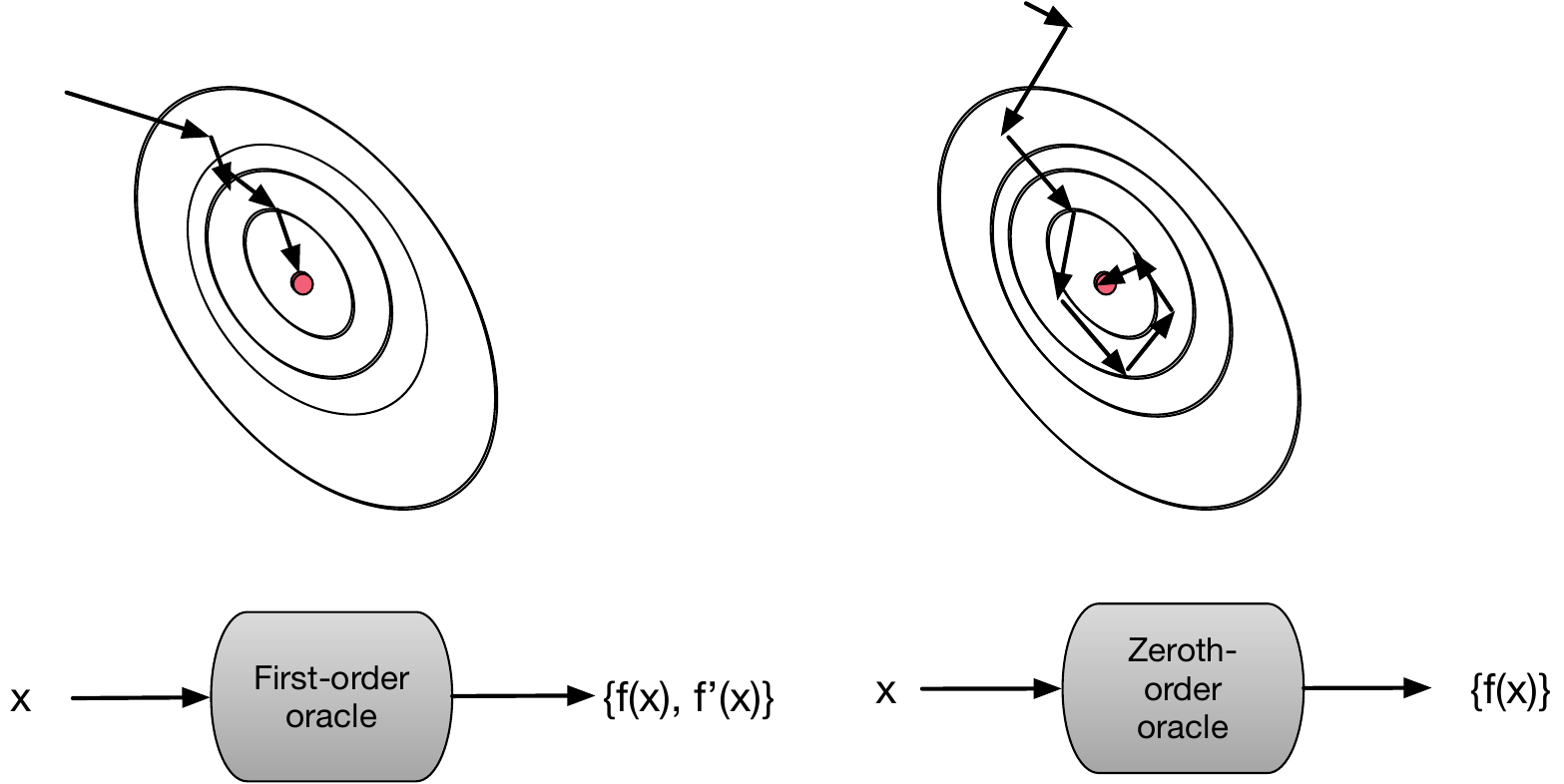}
    \caption{Zeroth-order optimization methods for generative AI are based on stochastic approximation, and average noisy values of the function to approximate gradient steps. Such methods define probabilistic coalgebras \cite{SOKOLOVA20115095}.}
    \label{zoodl}
\end{figure}

To illustrate how the broader coalgebraic definition of backpropagation is more useful than the previous definition in \cite{DBLP:conf/lics/FongST19}, we describe a class of generative AI methods based on adapting stochastic approximation \cite{rm} to deep learning, which are popularly referred to zeroth-order optimization \cite{liu2009large} (see Figure~\ref{zoodl}). A vast range of stochastic approximation methods have been explored in the literature (e.g., see \cite{borkar,kushner2003stochastic}). For example, in {\em random directions} stochastic approximation, each parameter is adjusted in a random direction by sampling from distribution, such as a multivariate normal, or a uniform distribution. Any of these zeroth-order stochastic approximation algorithms can itself be defined as a functor over symmetric monoidal categories

\[ L^{0}_{\epsilon}: {\tt Param} \rightarrow {\tt Learn}\]

where $\epsilon > 0$ is a real number defining a learning rate parameter that is gradually decayed. Notice now that the error of the approximation with respect to the target plays no role in the update process itself.  backpropagation, and $e(x,y): \mathbb{R} \times \mathbb{R} \rightarrow \mathbb{R}$ is a differentiable error function such that $\frac{\partial e}{\partial x}(x_0, -)$ is invertible for each $x_0 \in \mathbb{R}$. The functor $L^0_{\epsilon}$ defined by zeroth-order optimization sends each parameterized function $I: P \times A \rightarrow B$ to the learner $(P, I, U^0_I,r^0_I)$

\[ U^0_I(p, a, b) \coloneqq p - \epsilon I(p, a, b) \]

Here, the $1$-point gradient estimate is approximated by the (noisy) sampled value, averaged over multiple steps using a decaying learning rate as required by the convergence theorems of stochastic approximation \cite{rm,kushner2003stochastic}.  The advantages of zeroth-order stochastic approximation methods are that it avoids computing gradients over a very large number of parameters (which for state of the art generative AI models can be in the billions or trillions of parameters), and it potentially helps avoid local minima by stochastically moving around in a very high-dimensional space. The disadvantage is that it can be significantly slower than gradient-based methods for well-behaved (convex) functions.  

We can easily extend our previous definition of backpropagation as a coalgebra to capture zeroth-order optimization methods which act like stochastic dynamical systems,   where there is a distribution of possible ``next" states that is produced as a result of doing stochastic approximation updates. 

  \begin{definition}
{\bf Stochastic Backpropagation} defines an $F_{\mbox{SGD}}$-coalgebra over the symmetric monoidal category {\tt Param}, specified by an endofunctor $X \rightarrow F_{\mbox{SGD}}B(X)$ defined as 
\[ F_{\mbox{SGD}}(X) = A \times B \times {\cal D}(X)\]
  \end{definition}

  where $F_{\mbox{SGD}}$ defines the variant of backpropagation defined by stochastic gradient descent, and ${\cal D}$ is the distribution functor over $X$ that defines a distribution over possible objects $X$ in {\tt Param}. There is a vast literature on stochastic coalgebras that can be defined in terms of such distribution functors. \cite{SOKOLOVA20115095} contains an excellent review of this literature.

\section{Universal Properties of Coalgebras}

Category theory can be viewed as the study of {\em universal properties} \citep{riehl2017category}. To illustrate this abstract idea in the setting of URL, we can  build complex universal coalgebra systems out of simpler systems, focusing for now on $F$-coalgebras over the category of sets. All of the standard universal constructions in the category of sets straightforwardly carry over to universal coalgebras over sets. 

\begin{itemize}
    \item The {\em coproduct} (or sum) of two $F$-coalgebras $(S, \alpha_S)$ and $(T, \alpha_T)$ is defined as follows. Let $i_S: S \rightarrow (S + T)$ and $i_T: T \rightarrow (S + T)$ be the injections (or monomorphisms) of the sets $S$ and $T$, respectively, into their coproduct (or disjoint union) $S + T$. From the universal property of coproducts, defined in the previous Section, it follows that there is a unique function $\gamma: (S + T) \rightarrow F(S+T)$ such that $i_S$ and $i_T$ are both homomorphisms: 

    \begin{center}
\begin{tikzcd}
  S \arrow[r, "i_S"] \arrow[d, "\alpha_S"]
    & S+T \arrow[d, "\gamma" ] \\
  F(S) \arrow[r,  "F(i_S)"]
& F(S+T)
\end{tikzcd}

\begin{tikzcd}
  T \arrow[r, "i_T"] \arrow[d, "\alpha_T"]
    & S+T \arrow[d, "\gamma" ] \\
  F(R) \arrow[r,  "F(i_T)"]
& F(T)
\end{tikzcd}
\end{center}

\item To construct the {\em co-equalizer} of two homomorphisms $f: (S \alpha_S) \rightarrow (T, \alpha_T)$ and $g: (S, \alpha_S) \rightarrow (T, \alpha_T)$, we need to define a system $(U, \alpha_U)$ and a homomorphism $h: (T, \alpha_T) \rightarrow (U, \alpha_T)$ such that $h \circ f = h \circ g$, and for every homomorphism $h': (T, \alpha_T) \rightarrow (U', \alpha_{U'})$ such that $h' \circ f = h' \circ g$, there exists a unique homomorphism $l: (U, \alpha_U) \rightarrow (U', \alpha_{U'})$ with the property that $h'$ factors uniquely as $h' = l \circ h$. Since $f$ and $g$ are by assumption set-valued functions $f: S \rightarrow T$ and $g: S \rightarrow T$, we know that there exists a coequalizer $h: T \rightarrow U$ in the category of sets (by the property that coequalizers exist in {\bf Sets}). If we define $F(h) \circ \alpha_T: T \rightarrow F(U)$, then since

\begin{eqnarray*}
    F(h) \circ \alpha_T \circ f &=& F(h) \circ F(f) \circ \alpha_S \\
    &=& F(h \circ f) \circ \alpha_S \\
    &=& F(h \circ g) \alpha_S \\
    &=& F(h) \circ F(g) \circ \alpha_S \\
    &=& F(h) \circ \alpha_T \circ g
\end{eqnarray*}

and given $h: T \rightarrow U$ is a coequalizer, there must exist a unique function $\alpha_U: U \rightarrow F(U)$ with the property we desire.

\item There is a more general way to establish the existence of coproducts, coequalizers, pullbacks, and in general colimits by defining a {\em forgetful functor} $U: Set_F \rightarrow Set$, which sends a coalgebra system to its underlying set $U(S, \alpha_S) = S$, and sends the $F$-homomorphism $f: (S, \alpha_S) \rightarrow (T, \alpha_T)$ to the underlying set-valued function $f: S \rightarrow T$. We can apply a general result on the creation of colimits as follows: 

\begin{theorem}
    The forgetful functor $U: Set_F \rightarrow F$ creates colimits, which means that any type of colimit in $Set_F$ exists, and can be obtained by constructing the colimit in Set and then defining for it in a unique way an $F$-transition dynamics. 
\end{theorem}

\item For pullbacks and limits, the situation is a bit more subtle. If $F: Set \rightarrow Set$ preserves pullbacks, then pullbacks in Set$_F$ can be constructed as follows. Let $f: (S, \alpha_S) \rightarrow (S, \alpha_T)$ and $g: (S, \alpha_S) \rightarrow (T, \alpha_T)$ be homomorphisms. Define the pullback of $f$ and $g$ in Set as follows, and extend that to get a pullback of $F(f)$: 

   \begin{center}
\begin{tikzcd}
  P \arrow[r, "\pi_1"] \arrow[d, "\pi_2"]
    & S \arrow[d, "f" ] \\
  U \arrow[r,  "g"]
& T
\end{tikzcd}

\begin{tikzcd}
  F(P) \arrow[r, "F(\pi_1)"] \arrow[d, "F(\pi_2)"]
    & F(S) \arrow[d, "F(f)" ] \\
  F(U) \arrow[r,  "F(g)"]
& F(T)
\end{tikzcd}

\end{center}

It can be shown that the pullback of two homomorphisms is a bisimulation even in the case that $F$ only preserves {\em weak} pullbacks (recall that in defining the pullback universal construction, it was necessary to assert a uniqueness condition, which is weakened to an existence condition in weak pullbacks). 

\end{itemize}

\subsection{Lambek's Theorem and Final Coalgebras}

Finally, we illustrate how category theory gives rise to a novel notion of state, which generalizes the ideas explored in RL thus far.  The {\em final}  coalgebras defined by a functor that represents a monotone function over the category defined by a preorder $(S, \leq)$, where $S$ is a set and $\leq$ is a relation that is reflexive and transitive. That is, $a \leq a, \forall a \in S$, and if $a \leq b$, and $b \leq c$, for $a, b, c \in S$, then $a \leq c$. Note that we can consider $(S, \leq)$ as a category, where the objects are defined as the elements of $S$ and if $a \leq b$, then there is a unique arrow $a \rightarrow b$. 

Let us define a functor $F$ on a preordered set $(S, \leq)$ as any monotone mapping $F: S \rightarrow S$,  so that if $a \leq b$, then $F(a) \leq F(b)$. Now, we can define an $F$-algebra as any {\em pre-fixed point} $x \in S$ such that $F(x) \leq x$. Similarly, we can define any {\em post-fixed point} to be any $x \in S$ such that $x \leq F(x)$. Finally, we can define the {\em final $F$-coalgebra} to be the {\em greatest post-fixed point} $x \leq F(x)$, and analogously, the {\em initial $F$-algebra} to be the least pre-fixed point of $F$. 

In this section, we give a detailed overview of the concept of {\em final coalgebras} in the category of coalgebras parameterized by some endofunctor $F$. This fundamental notion plays a central role in the application of universal coalgebras to model a diverse range of AI and ML systems. Final coalgebras generalize the concept of (greatest) fixed points in many areas of application in AI, including causal inference, game theory and network economics, optimization, and reinforcement learning among others. The final coalgebra, simply put, is just the final object in the category of coalgebras. From the universal property of final objects, it follows that for any other object in the category, there must be a unique morphism to the final object. This simple property has significant consequences in applications of the coalgebraic formalism to AI and ML, as we will see throughout this paper. 

An $F$-system $(P, \pi)$ is termed {\bf final} if for another $F$-system $(S, \alpha_S)$, there exists a unique homomorphism $f_S: (S, \alpha_S) \rightarrow (P, \pi)$. That is, $(P, \pi)$ is the terminal object in the category of coalgebras $Set_F$ defined by some set-valued endofunctor $F$. Since the terminal object in a category is unique up to isomorphism, any two final systems must be isomorphic. 

\begin{definition} 
    An $F$-coalgebra $(A, \alpha)$ is a {\em fixed point} for $F$, written as $A \simeq F(A)$ if $\alpha$ is an isomorphism between $A$ and $F(A)$. That is, not only does there exist an arrow $A \rightarrow F(A)$ by virtue of the coalgebra $\alpha$, but there also exists its inverse $\alpha^{-1}: F(A) \rightarrow A$ such that 

    \[ \alpha \circ \alpha^{-1} = \mbox{{\bf id}}_{F(A)} \ \ \mbox{and} \ \  \alpha^{-1} \circ \alpha = \mbox{{\bf id}}_A \]
\end{definition}

The following lemma was shown by Lambek, and implies that the transition structure of a final coalgebra is an isomorphism. 

\begin{theorem}
    {\bf Lambek:} A final $F$-coalgebra is a fixed point of the endofunctor $F$. 
\end{theorem}

{\bf Proof:} The proof is worth including in this paper, as it provides a classic example of the power of diagram chasing. Let $(A, \alpha)$ be a final $F$-coalgebra. Since $(F(A), F(\alpha)$ is also an $F$-coalgebra, there exists a unique morphism $f: F(A) \rightarrow A$ such that the following diagram commutes: 
\begin{center}
\begin{tikzcd}
  F(A) \arrow[r, "f"] \arrow[d, "F(\alpha)"]
    & A \arrow[d, "\alpha" ] \\
  F(F(A)) \arrow[r,  "F(f)"]
& F(A)
\end{tikzcd}
\end{center} 
However, by the property of finality, the only arrow from $(A, \alpha)$ into itself is the identity. We know the following diagram also commutes, by virtue of the definition of coalgebra homomorphism:
\begin{center}
\begin{tikzcd}
  A \arrow[r, "\alpha"] \arrow[d, "\alpha"]
    & F(A) \arrow[d, "\alpha" ] \\
  F(A) \arrow[r,  "F(\alpha)"]
& F(F(A))
\end{tikzcd}
\end{center} 
Combining the above two diagrams, it clearly follows that $f \circ \alpha$ is the identity on object $A$, and it also follows that $F(\alpha) \circ F(f)$ is the identity on $F(A)$. Therefore, it follows that: 

\[ \alpha \circ f  = F(f) \circ F(\alpha) = F(f \circ \alpha) = F(\mbox{{\bf id}}_A) = \mbox{{\bf id}}_{F(A)} \qed \]

By reversing all the arrows in the above two commutative diagrams, we get the easy duality that the initial object in an $F$-algebra is also a fixed point. 

\begin{theorem}
    {\bf Dual to Lambek}: The initial $F$-algebra $(A, \alpha)$, where $\alpha: F(A) \rightarrow A$, in the category of $F$-algebras is a fixed point of $F$. 
\end{theorem}

The proof of the duality follows in the same way, based on the universal property that there is a unique morphism from the initial object in a category to any other object. 

Lambek's lemma  has many implications in URL, one of which is that final coalgebras generalize the concept of a (greatest) fixed point used previously to characterize dynamic programming methods in MDPs \citep{DBLP:books/lib/Bertsekas05}. 

\subsection{Asynchronous Distributed Computation in Universal Coalgebras} 

\label{urlcoalg} 

We can briefly summarize the main points of this paper by rephrasing Algorithm~\ref{adm} in the general coalgebraic setting as shown below. The problem of minimization of a vector function $F: X \rightarrow X$ is generalized to finding a final coalgebra with a specified $F$-dynamics, where $F$ is some functor in a symmetric monoidal category ${\cal C}$ with tensor product $\otimes$. 

\begin{algorithm}
\caption{A Asynchronous Distributed Algorithm for Finding Final Coalgebras}
\label{urlalgm}

{\bf Input:} Given a functor diagram $F: {\cal J} \rightarrow {\cal C}$, where ${\cal C}$ is a category of coalgebras $\alpha_f = X \rightarrow f(X)$ that is (co)complete, meaning that it has categorical (co)products, (co)equalizers, and more generally has all finite (co)limits. This property ensures that any diagram $F$ is ``solvable" by finding its limit $\lim_F \in {\cal C}$. To concretize this abstraction, note that Algorithm~\ref{adm} defined a product function $F$, which can be seen as the limit of the diagram ${\cal J} = \bullet \bullet \cdots \bullet$. For example, in Section~\ref{topos-url}, we showed the category ${\cal C}_Q$ of action-value functions defines a topos, so that the coalgebras over ${\cal C}_Q$ will in addition to having (co)limits, also have a subobject classifier and exponential objects as part of their diagrammatic vocabulary. 

{\bf Output:} A final coalgebra $X^* \rightarrow F(X^*)$  such that $X^* \cong F(X^*)$ (where $\cong$ is the isomorphism in ${\cal C}$. 

\begin{algorithmic}[1]
\REPEAT

\STATE At each time step $t \in T$, where $T = \{0, 1, \ldots, \}$, update some component coalgebra $\alpha_{f_c} = X \rightarrow f_c(X)$ using an asynchronous distributed coalgebraic iteration, where $c$ is an object in ${\cal J}$. Note that in Algorithm~\ref{adm}, each component function $f_i$ corresponds to an object $\bullet$ in the diagram ${\cal J} = \bullet \bullet \cdots \bullet$. 

\[ X_c \rightarrow f_c(X) \]
\label{step1url}

\STATE Each update  of $X$ is done in parallel by some ``processor" whose ``information field" is measurable (see Section~\ref{udm}). For example, in deep RL (Section~\ref{deeprl}, each component coalgebra is a neural network layer for which the errors at its output stage are known, and 

\[ X_c(t+1) = f(X_1(\tau^1_i(t)), \ldots X_{c_f}(\tau^i_n(t))),  \ \ \forall t \in T^i\]

where $c_f$ is the number of elements in the information field of element $c$. 

\STATE where $T^i$ is the set of time points where $X_f$ is updated, and $0 \leq \tau^i_j(t)  \leq t$, and at all times $t \notin T^i$, we assume that 

\[ X_f(t+1) = X_f(t) \]

\IF{the final coalgebra  $X^* \rightarrow F(X^*)$ is not found}

\STATE Set $t = t+1$, and return to Step~\ref{step1url}. 

\ELSE 

\STATE Set {\bf done} $\leftarrow$ {\bf true}. 

\ENDIF 

\UNTIL{{\bf done}}. 

\STATE  Return the final coalgebra $F(X^*) \cong X^*$ of $F$. 

\end{algorithmic}
\end{algorithm}

\section{Information Fields over Coalgebras} 
\label{udm} 

The essence of asynchronous distributed computation in both RL and URL is to update each component coalgebra when it is possible to do so. For example, in Algorithm~\ref{adm}, each processor updates the component function $f_i: X \rightarrow X$ sufficiently often so that the entire product function $F$ converges. Applied to $Q$-learning, we can think of each product function $f_i$ as the simply $Q(x,a)$, the state-action value. In general, if we specify an approximation architecture by a functor diagram, as in deep RL described in Section~\ref{deeprl}, each component coalgebra (e.g., a layer of a neural network, or a computing node) may not be ready to be updated until the information it needs is accessible. For deep RL, it means that backpropagation requires unrolling the updates from the last layer backwards.  More general functor diagrams may require other approaches. A very general theory of how to structure decision making in complex multi-agent settings was developed by Witsenhausen \citep{witsenhausen:1975}, which we generalized to the categorical setting in \citep{sm:udm}. 

We now proceed to give a brief introduction to the Universal Decision Model (UDM)\cite{sm:udm}. In the UDM category ${\cal C}_{\mbox{UDM}}$, as in any category, we are given a collection of {\em decision objects} ${\cal D}$, and a set of morphisms ${\cal M}_{\mbox{UDM}}$ between UDM objects, where $f: c \rightarrow d$ is a morphism that maps from UDM object $c$ to $d$. A morphism need not exist between every pair of UDM objects. In this paper, we restrict ourselves to {\em locally small} UDM categories, meaning that exists only a set's worth of morphisms between any pair of UDM objects. 
 
 \begin{definition}  
 \label{udm-defn}
 A Universal Decision Model (UDM)  is defined as a category ${\cal C}_{\mbox{UDM}}$, where each decision object is represented as a  tuple $\langle (A, (\Omega, {\cal B}, P), U_\alpha, {\cal F}_\alpha, {\cal I}_\alpha)_{\alpha \in A} \rangle$, where $A$ in URL represent coalgebras,  $(\Omega, {\cal B}, P)$ is a probability space representing the inherent stochastic state of nature due to randomness,  $U_\alpha$ is a measurable space from which a decision $u \in U_\alpha$ is chosen by decision object $\alpha$. Each element's policy in a  decision object is any function $\pi_\alpha: \prod_\beta U_\beta \rightarrow U_\alpha$ that is measurable from its information field ${\cal I}_\alpha$, a subfield of  the overall product space $(\prod_\alpha U_\alpha, \prod_\alpha {\cal F}_\alpha)$, to the $\sigma$-algebra ${\cal F}_\alpha$. The policy of decision object $\alpha$ can be any function $\pi_\alpha: \prod_\beta U_\beta \rightarrow U_\alpha$. 
\end{definition} 

\begin{definition}  
The {\bf  information field} of an element $\alpha \in A$ in a decision object $c$ in UDM category ${\cal C}_{\mbox{UDM}}$ is denoted as ${\cal I}_\alpha \subset {\cal F}_A(A)$ characterizes the information available to decision object $\alpha$ for choosing a decision $u \in U_\alpha$. 
\end{definition}  

To ground this definition out in terms of the stochastic approximation theory of $Q$-learning, an information field precisely delineates what information is available to each of the parallel asynchronous distributed processors that are updating the $Q$-function. 
The information field structure yields a surprisingly rich topological space that has many important consequences for how to organize the  decision makers in a complex organization into subsystems. An element $\alpha$ in a decision object requires information from other elements or subsystems in the network. To formalize this notion, we use product decision fields and product $\sigma$-algebras, with their canonical projections.

\begin{definition}  
  Given a subset of nodes $B \subset A$, let $H_B = \Omega \times \prod_{\alpha \in B} U_\alpha$ be the {\bf product space of decisions} of nodes in the subset $B$, where the {\bf product $\sigma$-algebra} is ${\cal B} \times \prod_{\alpha \in B} {\cal F}_\alpha = {\cal F}_B(B)$. It is common to also denote the product $\sigma$-algebra by the notation $\otimes_{\alpha \in A} {\cal F}_\alpha$.  If $C \subset B$, then the {\bf induced $\sigma$-algebra} ${\cal F}_B(C)$ is a subfield of ${\cal F}_B(B)$, which can also be viewed as the inverse image of ${\cal F}_C(C)$ under the canonical projection of $H_B$ onto $H_C$. \footnote{Note that for any cartesian product of sets $\prod_i X_i$, we are always able to uniquely define  a projection map into any component set $X_i$, which is a special case of the product universal property in a category.} 
 \end{definition} 
 
 \subsection{Bisimulation Morphisms \index{bisimulation morphism} as Open Maps}

In a UDM category, the morphisms between decision objects are represented using the concept of {\em open maps}, as proposed in \cite{DBLP:conf/lics/JoyalNW93}. This framework is based on defining a model of computation as a category.  A related notion has been proposed for probabilistic bisimulation \cite{SOKOLOVA20115095}.   One fundamental notion is {\em bisimulation} between machines or processes using {\em open maps} in categories \cite{DBLP:conf/lics/JoyalNW93}. This definition can be seen as a generalization of the simpler bisimulation relationship that exists for the category of labeled transition systems \cite{DBLP:conf/lics/JoyalNW93}, which are specified as a relation of tuples $(s, a, s')$, which  indicates a transition from state $s$ to state $s'$, where $s, s' \in S$, and $a \in A$.  Given a collection of labeled transition systems, each of which is represented as an object, morphisms are defined from one object to another that preserve the dynamics under the labeling function. For example, a surjective function $f: S \rightarrow S'$ maps states in object $X$ to corresponding states in $Y$, where the labels are mapped as well, with the proviso that some transitions in $X$ may be hidden in $Y$ (i.e., cause no transition). If a morphism exists between objects $X$ and $Y$, then $Y$ is said to be a bisimulation of $X$.

Let ${\cal M}$ denote a model of computation, where a morphism $m: X \rightarrow Y$ is to intuitively viewed as a simulation of $X$ in $Y$. Within ${\cal M}$, we choose a subcategory of ``observation objects" and ``observation extension" morphisms between them. We can denote this category of observations by ${\cal P}$. Given an observation object $P \in {\cal P}$, and a model $X \in {\cal M}$, $P$ is said to be an {\em observable behavior} of $X$ if there is a morphism $o: P \rightarrow X$ in ${\cal M}$. We define morphisms $m: X \rightarrow Y$ that have the property that whenever an observable behavior of $X$ can be extended via $f$ in $Y$, that extension can be matched by an extension of the observable behavior in $X$. 

\begin{definition} \cite{DBLP:conf/lics/JoyalNW93}
\label{popen}
A morphism $m: X \rightarrow Y$ in a model of computation ${\cal M}$ is said to be ${\cal P}$-open if whenever $f: O_1 \rightarrow O_2$ in ${\cal P}$, $p: O_2 \rightarrow Y$ in ${\cal M}$, and $q: O_2 \rightarrow Y$ in ${\cal M}$, the below  diagram commutes, that is, $m \ p = q \ f$. 
\begin{center}
\begin{tikzcd}
  O_1 \arrow{d}{f} \arrow{r}{p}
    & X \arrow[red]{d}{m} \\
  O_2 \arrow[red]{r}[blue]{q}
&Y \end{tikzcd}
\end{center} 
\end{definition} 
This definition means that whenever such a ``square" in ${\cal M}$ commutes, the path $f \ p$ in $Y$ can be extended via $m$ to a path $q$ in $Y$, there is a ``zig-zag" mediating morphism $p'$ such that the two triangles in the diagram below
\begin{center}
\begin{tikzcd}
  O_1 \arrow{d}{f} \arrow{r}{p}
    & X \arrow[red]{d}{m} \\
  O_2  \arrow[red]{ur}[blue]{p'} \arrow[red]{r}[blue]{q}
&Y \end{tikzcd}
\end{center} 
commute, namely $p = p' \ f$ and $q = m \ p'$. We now define the abstract definition of {\em bisimulation} as follows: 

\begin{definition} 
Two models $X$ and $Y$ in ${\cal M}$ are said to be ${\cal P}$-bisimilar (in ${\cal M}$) if there exists a span of open maps from a common object $Z$: 
\begin{center}
 \begin{tikzcd}[column sep=small]
& Z \arrow{dl}{m} \arrow{dr}{m'} & \\
  X  &                         & Y
\end{tikzcd}
\end{center} 
\end{definition} 

Note that if the category ${\cal M}$ has pullbacks, then the $\sim_{{\cal P}}$ is an equivalence relation, which induces a quotient mapping. Furthermore, pullbacks of open map bisimulation mappings are themselves bisimulation mappings. Many of the bisimulation mappings studied for MDPs and PSRs are special cases of the more general formalism above. 

\subsection{Bisimulation in UDMs}

We introduce the concept of bisimulation morphisms \index{bisimulation morphism} between UDM objects, which builds on a longstanding theme in computer science on using category theory to understand machine behavior. One fundamental notion is {\em bisimulation} between machines or processes using {\em open maps} in categories \cite{DBLP:conf/lics/JoyalNW93}. 

\begin{definition} 
\label{bisim-defn}
The {\bf bisimulation} relationship between two UDM objects $M = \langle A, (\Omega, {\cal B}, (U_\alpha, {\cal F}_\alpha, {\cal I}_\alpha)_{\alpha \in A}\rangle$ and $M' = \langle A', (\Omega', {\cal B}', (U'_\alpha, {\cal F}'_\alpha, {\cal I}'_\alpha)_{\alpha \in A'}\rangle$,  denoted as $M \twoheadrightarrow M'$, is defined as  is defined by a tuple of surjections as follows: 
\begin{itemize}
    \item A surjection $f: A \twoheadrightarrow A'$ that maps  elements in $A$ to corresponding elements in $A'$. As $f$ is surjective, it induces an equivalence class in $A$ such that $x \sim y, x, y \in A$ if and only if $f(x) = f(y)$. 
    \item A surjection $g: H \twoheadrightarrow H'$, where $H = \Omega \times \prod_{\alpha \in A} U_\alpha$, with the product $\sigma$-algebra ${\cal B} \times \prod_{\alpha \in A} {\cal F}_\alpha$, and $H' = \Omega' \times \prod_{\alpha \in A'} U'_\alpha$, with the corresponding $\sigma$-algebra ${\cal B}' \times \prod_{\alpha \in A'} {\cal F}'_\alpha$. 
\end{itemize}
\end{definition} 

This definition can be seen as a generalization of the simpler bisimulation relationship that exists for the category of labeled transition systems \cite{DBLP:conf/lics/JoyalNW93}, which are specified as a relation of tuples $(s, a, s')$, which  indicates a transition from state $s$ to state $s'$, where $s, s' \in S$, and $a \in A$.  Given a collection of labeled transition systems, each of which is represented as an object, morphisms are defined from one object to another that preserve the dynamics under the labeling function. For example, a surjective function $f: S \rightarrow S'$ maps states in object $X$ to corresponding states in $Y$, where the labels are mapped as well, with the proviso that some transitions in $X$ may be hidden in $Y$ (i.e., cause no transition). If a morphism exists between objects $X$ and $Y$, then $Y$ is said to be a bisimulation of $X$. The state-dependent action recoding in the MDP homomorphism definition is captured by the equivalent surjection $g$ that maps the product space $H$ with its associated $\sigma$-algebra to $H'$ with its corresponding $\sigma$-algebra. 

Note that the general framework of bisimulations applied to universal coalgebras also applies here, and in fact, we could rephrase the above definitions in terms of universal coalgebras as well. We can specialize the definition in a number of ways, depending on the exact form chosen for the surjection $g$ between product spaces $H$ and $H'$. Since the surjection $f$ maps decision makers into equivalence classes, each decision maker in model $M'$ corresponds to an equivalence class of decision makers in model $M$. Thus, we need to collapse their corresponding information fields. We can define the information field of an equivalence class of agents $[\alpha]_f$, meaning all $\beta$ such that $f(\beta) = f(\alpha)$, by recalling that an information field is a subfield of the product field, and as it is a lattice, we can use the join operation, as defined below: 

\begin{definition} 
The {\bf quotient information field} of a collection of agents $[\alpha]_f$ is defined as the join of the information fields of each agent: 
\begin{equation}
\label{qif}
{\cal I}_{[\alpha]} = \bigvee_{\beta \in [\alpha]_f} {\cal I}_\alpha
\end{equation}
\end{definition} 

\subsection{Observation Objects in UDM}

We now briefly discuss observation objects in a UDM. Observation objects, as mentioned above, represent observable trace behavior of a decision object. We first define observation functions that underlie information fields.

\begin{definition} 
For a UDM object $M = \langle A, (\Omega, {\cal B}, (U_\alpha, {\cal F}_\alpha, {\cal I}_\alpha)_{\alpha \in A} \rangle$ over a finite $\sigma$-algebra, the observations $Z_1, \ldots, Z_k$ taking values in a measurable space $(Z_i, {\cal Z}_i)$, where $Z_i = \eta_i(\omega, U_1, \ldots, U_{|A|})$ is an {\bf observation generation map} function  such that $\sigma(Z_{1:k})$ is the smallest $\sigma$-algebra contained in ${\cal B} \otimes \prod_\alpha {\cal F}_\alpha$ with respect to which the observation maps $\eta_i$ are measurable functions. We say the observations $Z_1, \ldots, Z_k$ generate the information field ${\cal I}_\alpha$ if 
\begin{equation}
    \sigma(Z_1, \ldots, Z_k) = {\cal I}_\alpha
\end{equation}
\end{definition} 

We can then define an observation object associated with a UDM decision object as one equipped with an observation generation map that can generate the various information fields in the decision object. 

\begin{definition} 
A UDM {\bf observation object} $O = \langle A, (\Omega, {\cal B}, (U_\alpha, {\cal F}_\alpha, \eta_\alpha)_\alpha \rangle$ is such that each information field ${\cal I}_\alpha$ can be generated from the associated observation generation map $\eta_\alpha$. 
\end{definition} 

We can define an observation morphism between an observation object $O$ and a decision object $M$ to be one such that $O$ represents an observable behavior of $M$, and extend the notion of ${\cal P}$-open morphisms from Definition~\ref{popen} above.

\subsection{Solvability of UDM objects \index{Causality in UDM} } 

Each decision maker $\alpha$  in a UDM object  has associated with it a control law or policy $\pi_\alpha: H \rightarrow U_\alpha$, which is measurable from its information field ${\cal I}_\alpha$ to ${\cal F}_\alpha$, the measurable space associated with its decision space $U_\alpha$. Essentially, this means that any pre-image of $\pi^{-1}_\alpha(E)$, for any measurable subset $E \subset {\cal F}_\alpha$, is also  measurable on its information field, that is $\pi^{-1}_\alpha(E) \subset {\cal I}_\alpha$. For any subsystem in the cloud computing network, the overall policy space $\pi_B = \prod_{\alpha \in B} \pi_alpha$ is given by the product space of all individual control laws. 

\begin{definition}  
\label{solvable}
A UDM object $\langle A, (\Omega, {\cal B}, P), (U_\alpha, {\cal F}_\alpha, {\cal I}_\alpha)_{\alpha \in A} \rangle$ is said to be {\bf solvable} if for every state of nature $\omega \in \Omega$, and every control law $\pi \in \Pi_A$, the set of simultaneous equations given below has one and only one solution $u \in U$. 
\begin{equation}
    u_\alpha = \pi_\alpha(h) \equiv \pi_\alpha(\omega, u)
\end{equation}
Here, $\pi_\alpha$ can be viewed as a projection from the joint decision $h$ taken by the entire ensemble of decision makers in the intrinsic model. A UDM category ${\cal C}_{\mbox{UDM}}$ is solvable if every object in it is solvable. 
\end{definition}  

Intuitively, the solvability criterion states that a UDM object represents a solvable decision problem if each agent in the object can successfully compute its response, given access to its information field, and that its response is uniquely determined for every state of nature. It is easy to construct unsolvable decision objects. Consider a simple network with two elements $\alpha$ and $\beta$, each of whose information fields includes the measurable space of the other. In this case, neither element can compute its function without knowing the other's response, hence both are waiting for the other to compute their response, and a deadlock ensures. 

Given the notion of solvability above, we can now define solution objects in a UDM. 

\begin{definition}  
\label{solution}
A UDM {\bf solution object} $\langle A, (\Omega, {\cal B}, P), (U_\alpha, \pi_\alpha, {\cal F}_\alpha, {\cal I}_\alpha)_{\alpha \in A} \rangle$ is defined as one for which for every state of nature $\omega \in \Omega$, the control law $\pi_\alpha$ uniquely defines a fixed point solution $u_\alpha = \pi_\alpha(h) \equiv \pi_\alpha(\omega, u)$ to the associated decision object. 
\end{definition}  

We can straightforwardly define morphisms between solution objects and decision objects. 
To understand the causality condition, it is crucial to organize the decision makers into a partial order, such that for every total ordering that can be constructed from the partial ordering, the agents can successfully compute their functions based on the computations of agents that preceded them in the ordering. 

\begin{definition}  
\label{causal}
A UDM object  $\langle A, (\Omega, {\cal B}, P), (U_\alpha, {\cal F}_\alpha, {\cal I}_\alpha) \rangle$ is said to be {\bf causal} if there exists at least one function $\phi: H \rightarrow S$, where $S$ is the set of total orderings of computing elements in $A$, satisfying the property that for any partial stage of the computation $1 \leq k \leq n$, and any ordered set $(\alpha_1, \ldots, \alpha_k)$ of distinct elements from $A$, the set $E \subset H$ on which $\phi(h)$ begins with the same ordering $(\alpha_1, \ldots, \alpha_k)$ satisfies the following causality condition: \begin{equation} 
\label{causality} 
\forall F \in {\cal F}_{\alpha_k},\ \ E \cap F \in {\cal F}(\{\alpha_1, \ldots, \alpha_{k-1} \})
\end{equation} 
\end{definition}  
In other words, if at every step of the process, the $k^{\mbox{th}}$ decision making element $\alpha_k$ can successfully compute its response based on the information fields of the past $k-1$ elements, the system is then considered causal. These definitions are what is missing from previous analyses of stochastic approximation in asynchronous distributed models, and we can use these to more fully develop a theory of asynchronous iterative computation of final coalgebras in universal coalgebras. 

\section{The Metric Yoneda Lemma: Coinductive Inference in Generalized Metric Spaces}
\label{metricyoneda}

In this section, we want to use a specific form of the Yoneda Lemma, called the {\em metric Yoneda Lemma} \citep{BONSANGUE19981} to derive a form of coinductive inference in generalized metric spaces. These are based on a generalized notion of distance that combines familiar spaces, such as preorders and usual metric spaces, where the typical axioms of a distance metric are relaxed. The advantage of the metric Yoneda Lemma is that it can be shown that notions such as completion of a metric space arise from the metric Yoneda Lemma. 

\begin{figure}[t]
\begin{minipage}{0.75\textwidth}
\includegraphics[scale=0.8]{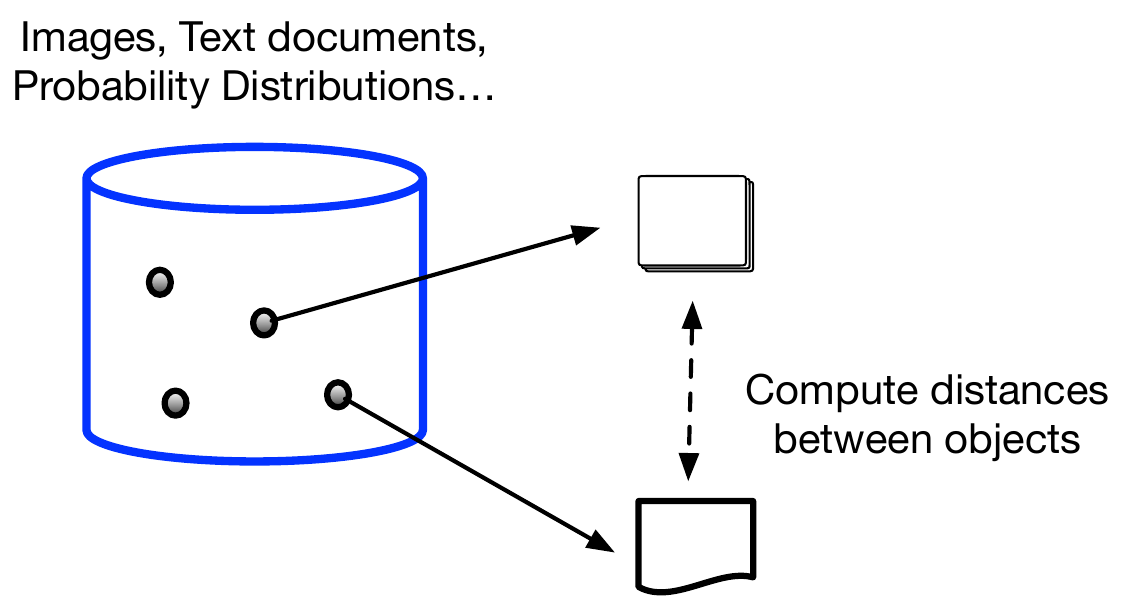}
\end{minipage}
\caption{The convergence of many algorithms in RL is usually analyzed using distances between objects in a {\em metric space}. Interpreting distances categorically leads to powerful ways to reason about convergence of RL in generalized metric spaces.}   
\label{metric-space}
\end{figure}

Figure~\ref{metric-space} illustrates a common motif in analyzing the convergence of many RL algorithms \citep{tsitsiklis,borkar}: define the problem in terms of computing distances between a group of objects in a (generalized) metric space. Examples of objects include points in $n$-dimensional Euclidean space, probability distributions, text documents represented as strings of tokens, and images represented as matrices.

\subsection{Generalized Metric Spaces as Categories}

\begin{definition}
A {\em generalized metric space} $(X, d)$ is a set $X$ of objects, and a non-negative function $X(-,-): X \times X \rightarrow [0, \infty]$ that satisfies the following properties: 

\begin{enumerate}
    \item $X(x,x) = 0$: distance between the same object and itself is $0$. 

    \item $X(x,z) \leq X(x,y) + X(y,z)$: the famous {\em triangle inequality} posits that the distance between two objects cannot exceed the sum of distances between each of them and some other intermediate third object.
\end{enumerate}
\end{definition}

In particular, generalized metric spaces are not required to be {\em symmetric}, or satisfy the property that if the distance between two objects $x$ and $y$ is $0$ implies $x$ must be identical to $y$, or finally that distances must be finite. These additional three properties listed below are what defines the usual notion of a {\em metric} space: 

\begin{enumerate}
    \item If $X(x, y) = 0$ and $X(y, x) = 0$ then $x = y$. 

\item $X(x, y) = X(y, x)$. 

\item $X(x, y) < \infty$. 
\end{enumerate}

Here are some examples of generalized metric spaces: 

\begin{enumerate}
    \item Any preorder $(P, \leq)$ such that all $p, q, r \in P$, if $p \leq q$ and $q \leq r$, then, $p \leq r$, and $p \leq p$, where 

   \[ P(p,q) =     \left\{ \begin{array}{rcl}
         0 & \mbox{if}
         & p \leq q \\ \infty  & \mbox{if} & p \not \leq q
                \end{array}\right\} \] 

                 \item The set of strings $\Sigma^*$ over some alphabet defined as the set $\Sigma$ where the distance between two strings $u$ and $v$ is defined as 

   \[ \Sigma^*(u,v) =     \left\{ \begin{array}{rcl}
         0 & \mbox{if}
         & u \ \mbox{is a prefix of} \ v \\ 2^{-n} & \mbox{otherwise} & \mbox{where} \ n \ \mbox{is the longest common prefix of } \ u \ \mbox{and} \ v
                \end{array}\right\} \] 

                 \item The set of non-negative distances $[0,\infty]$ where the distance between two objects $u$ and $v$ is defined as 

   \[ [0,\infty](u,v) =     \left\{ \begin{array}{rcl}
         0 & \mbox{if}
         & u \geq  v \\ v - u & \mbox{otherwise} & \mbox{where} \ r < s 
                \end{array}\right\} \] 

                 \item The powerset ${\cal P}(X)$ of all subsets of a standard metric space, where the distance between two subsets $V, W \subseteq X$ is defined as

   \[  {\cal P}(X)(V, W) = \inf \{ \epsilon > 0 | \forall v \in V, \exists w \in W, X(v, w) \leq \epsilon \} \] 

which is often referred to as the {\em non-symmetric Hausdorff distance}. 
                
\end{enumerate}

Generalized metric spaces can be shown to be $[0, \infty]$-enriched categories as the collection of all morphisms between any two objects itself defines a category.  In particular, the category $[0,\infty]$ is a complete and co-complete symmetric monoidal category. It is a category because objects are the non-negative real numbers, including $\infty$, and for two objects $r$ and $s$ in $[0,\infty]$, there is an arrow from $r$ to $s$ if and only if $r \leq s$. It is complete and co-complete because all equalizers and co-equalizers exist as there is at most one arrow between any two objects. The categorical product $r \sqcap s$ of two objects $r$ and $s$ is simply $\max\{r,s\}$, and the categorical coproduct $r \sqcup s$  is simply $\min\{r,s\}$. More generally, products are defined by supremums, and coproducts are defined by infimums. Finally, the {\em monoidal } structure is induced by defining the tensoring of two objects through ``addition":   

\[ +: [0, \infty] \times [0, \infty] \rightarrow [0,\infty]\]

where $r + s$ is simply their sum, and where as usual $r + \infty = \infty + r = \infty$. 

The category $[0,\infty]$ is also a {\em compact closed} category, which turns out to be a fundamentally important property, and can be simply explained in this case as follows. We can define an ``internal hom functor" $[0,\infty](-, -)$ between any two objects $r$ and $s$ in $[0, \infty]$ the distance $[0,\infty]$ as defined above, and the {\em co pre-sheaf} $[0,\infty](t,-)$ is {\em right adjoint} to $t + -$ for any $t \in [0, \infty]$. 

\begin{theorem}
For all $r, s$ and $t \in [0,\infty]$, 

\[ t + s \geq r \ \ \ \mbox{if and only if} \  \  \ s \geq [0,\infty](t,r)\]
\end{theorem}

We will explain the significance of compact closed categories for reasoning about AI and ML systems in more detail later, but in particular, we note that reasoning about feedback requires using compact closed categories to represent ``dual" objects that are diagrammatically represented by arrows that run in the ``reverse" direction from right to left (in addition to the usual convention of information flowing from left to right from inputs to outputs in any process model). 

We can also define a category of generalized metric spaces, where each generalized metric space itself as an object, and for the morphism between generalized metric spaces $X$ and $Y$, we can choose a {\em non-expansive function} $f: X \rightarrow Y$ which has the {\em contraction property}, namely 

\[ Y(f(x), f(y)) \leq c \cdot X(x,y) \]

where $0 < c < 1$ is assumed to be some real number that lies in the unit interval. The category of generalized metric spaces will turn out to be of crucial importance in this book as we will use a central result in category theory -- the Yoneda Lemma -- to give a new interpretation to distances. 

\subsection{The Metric Yoneda Lemma}

Finally, let us state a ``metric" version of the Yoneda Lemma specifically for the case of $[0,\infty]$-enriched categories in generalized metric spaces: 

\begin{theorem}
({\bf Yoneda Lemma for generalized metric spaces}): Let $X$ be a generalized metric space. For any $x \in X$, let 

\[ X(-, x): X^{\mbox{op}} \rightarrow [0, \infty], \ \ y \longmapsto X(y, x)\]
\end{theorem}

Intuitively, what the generalized metric version of the Yoneda Lemma is stating is that it is possible to represent an element of a generalized metric space by its co-presheaf, exactly analogous to what we saw above in the previous section for causal inference! If we use the notation

\[ \hat{X} = [0, \infty]^{X^{\mbox{op}}}\]

to indicate the set of all non-expansive functions from $X^{\mbox{op}}$ to $[0, \infty]$, then the Yoneda embedding defined by $y \longmapsto X(y, x)$ is in fact a non-expansive function, and itself an element of $\hat{X}$! Thus, it follows from the general Yoneda Lemma that for any other element $\phi$ in $\hat{X}$, 

\[ \hat{X}(X(-, x), \phi) = \phi(x) \]

Another fundamental result is that the Yoneda embedding for generalized metric spaces is an {\em isometry}. Again, this is exactly analogous to what we saw above for causal inference, which we denoted as the causal reproducing property. 

\begin{theorem}
The Yoneda embedding $y: X \rightarrow \hat{X}$, defined for $x \in X$ by $y(x) = X(-, x)$ is {\em isometric}, that is, for all $x, x' \in X$, we have: 

\[ X(x, x') = \hat{X}(y(x), y(x')) = \hat{X}(X(-, x), X(-, x'))\]
\end{theorem}

\subsection{Applying the Metric Yoneda Lemma in URL}

Let us briefly explain the significance of the metric Yoneda Lemma in URL. As we described earlier in Section~\ref{sec:intro}, the principle of metric coinduction states that any eventually contractive operator has a fixed point, namely a final coalgebra. Crucial to that theorem is the notion of contraction, which was specified in regular metric spaces. The metric Yoneda Lemma shows us how to generalize previous work to coinduction in generalized metric spaces, because it tells us how to construct isometric embeddings. We can apply the metric Yoneda Lemma to analyze Algorithm~\ref{urlalgm} in generalized metric spaces, which we postpone to future work. 

 \section{Topos Theory in URL: Complete Categories with Exponentials and Subobjects}
\label{urltopos} 

We now generalize the product construction in Algorithm~\ref{adm} to categories where we have not only products, but also (co)limits, exponential objects, and subobject classifiers. These constructions will give us a rich language to specify diagrams, which can then provide recipes for building URL algorithms and architectures. 

To concretize these abstractions,  let us explore defining a category over all value functions ${\cal C_V}$. A topos \citep{Johnstone:topostheory} is a ``set-like" category which generalizes all common operations on sets. Thus, the generalization  of subset is a subobject classifier in a topos.  To help build some intuition, consider how to define subsets without ``looking inside" a set. Essentially, a subset $S$ of some larger set $T$ can be viewed as a ``monic arrow", that is, an injective (or 1-1) function $f: S \hookrightarrow T$. We first introduce some basic categorial concepts, including diagrams as functors, and universal constructions, such as (co)limits. 

\subsection{Diagrams in URL}

We introduce the concept of diagrams in URL.  A {\em diagram} of {\em shape} {\cal J} over category ${\cal C}$  is defined as the functor $F: {\cal J} \rightarrow {\cal C}$. In the typical case where RL involves a sequence of decision steps, the indexing  category ${\cal J}$ might be defined as $\bullet \rightarrow \bullet \rightarrow \bullet \ldots$, and where the category ${\cal C}$ could be one of the RL categories illustrated in Section~\ref{urlcat}. Here are some canonical diagrams that carry special meaning when interpreted in terms of MDP or PSR homomorphisms. You can interpret each $\bullet$ is mapped into an MDP or PSR, and each arrow as an MDP or PSR homomorphism. 

\begin{enumerate}
    \item {\bf Pullback diagram}: The pullback diagram is defined as ${\cal J} = \bullet \rightarrow \bullet \leftarrow \bullet$.

    \item {\bf Pushforward diagram}: The pushforward diagram is defined as ${\cal J} = \bullet \leftarrow \bullet \rightarrow \bullet$. 

    \item {\bf Equalizer diagram}: The equalizer diagram is defined as  ${\cal J} = \bullet \rightarrow \bullet \twoarrows[0.4ex]{\rightarrow}{\rightarrow}  \bullet$. 

     \item {\bf Co-Equalizer diagram}: The co-equalizer diagram is defined as ${\cal J} = \bullet \twoarrows[0.4ex]{\rightarrow}{\rightarrow}  \bullet \rightarrow \bullet$. 
     
\end{enumerate}

In general, it can be shown that any diagram can be built as a combination of such elementary building blocks. Of course, it remains to be seen whether these diagrams are actually ``solvable". We will prove that in many cases of interest, all these diagrams are solvable, e.g., such as in the category of sets, ${\cal C}_{\bf Sets}$, as well as in other categories, such as topological spaces, groups etc.  To understand how to solve a diagram,  we need to introduce the concept of (co)limits and universal properties.  That 
requires defining {\em natural transformations}.

\subsection{Limits and Colimits} 

For any object $c \in {\cal C}$ and any diagram of shape ${\cal J}$, the {\em constant functor} $c: {\cal J} \rightarrow {\cal C}$ maps every object $j$ of ${\cal J}$ to $c$ and every morphism $f$ in ${\cal J}$ to the identity morphisms $1_c$. We can define a constant functor embedding as the collection of constant functors $\Delta: C \rightarrow {\cal C}^{\cal J}$ that send each object $c$ in ${\cal C}$ to the constant functor at $c$ and each morphism $f: c \rightarrow c'$ to the constant natural transformation, that is, the natural transformation whose every component is defined to be the morphism $f$. 

\begin{definition} 
    A {\bf cone over} a diagram $F: {\cal J} \rightarrow {\cal C}$ with the {\bf summit} or {\bf apex} $c \in {\cal C}$ is a natural transformation $\lambda: c \Rightarrow F$ whose domain is the constant functor at $c$. The components $(\lambda_j: c \rightarrow Fj)_{j \in {\cal J}}$ of the natural transformation can be viewed as its {\bf legs}. Dually, a {\bf cone under} $F$ with {\bf nadir} $c$ is a natural transformation $\lambda: F \Rightarrow c$ whose legs are the components $(\lambda_j: F_j \rightarrow c)_{j \in {\cal J}}$.

    % https://q.uiver.app/#q=WzAsMyxbMiwwLCJjIl0sWzAsMiwiRiBqIl0sWzQsMiwiRmsiXSxbMCwxLCJcXGxhbWJkYV9qIl0sWzAsMiwiXFxsYW1iZGFfayIsMl0sWzEsMiwiRiBmIl1d
\[\begin{tikzcd}
	&& c \\
	\\
	{F j} &&&& Fk
	\arrow["{\lambda_j}", from=1-3, to=3-1]
	\arrow["{\lambda_k}"', from=1-3, to=3-5]
	\arrow["{F f}", from=3-1, to=3-5]
\end{tikzcd}\]
    
\end{definition} 
Cones under a diagram are referred to usually as {\em cocones}. Using the concept of cones and cocones, we can now formally define the concept of limits and colimits more precisely. 
\begin{definition} 
    For any diagram $F: {\cal J} \rightarrow {\cal C}$, there is a functor $\mbox{Cone}(-, F): {\cal C}^{op} \rightarrow \mbox{{\bf Set}}$, which sends $c \in {\cal C}$ to the set of cones over $F$ with apex $c$. Using the Yoneda Lemma (see Supplementary Materials),  a {\bf limit} of $F$ is defined as an object $\lim F \in {\cal C}$ together with a natural transformation $\lambda: \lim F \rightarrow F$, which can be called the {\bf universal cone} defining the natural isomorphism ${\cal C}(-, \lim F) \simeq \mbox{Cone}(-, F)$.  Dually, for colimits, we can define a functor $\mbox{Cone}(F, -): {\cal C} \rightarrow \mbox{{\bf Set}}$ that maps object $c \in {\cal C}$ to the set of cones under $F$ with nadir $c$. A {\bf colimit} of $F$ is a representation for $\mbox{Cone}(F, -)$. Once again, using the Yoneda Lemma, a colimit is defined by an object $\mbox{Colim} F \in {\cal C}$ together with a natural transformation $\lambda: F \rightarrow \mbox{colim} F$, which defines the {\bf colimit cone} as the natural isomorphism $C(\mbox{colim} F, -) \simeq \mbox{Cone}(F, -)$. 
\end{definition} 

\begin{figure}[h]
\centering
\begin{tikzcd}
  T
  \arrow[drr, bend left, "x"]
  \arrow[ddr, bend right, "y"]
  \arrow[dr, dotted, "k" description] & & \\
    & 
    U\arrow[r, "g'"] \arrow[d, "f'"]
      & X \arrow[d, "f"] \\
& Y \arrow[r, "g"] &Z
\end{tikzcd}
\caption{
 Universal property of pullback mappings. } 
\label{pullback}
\end{figure}
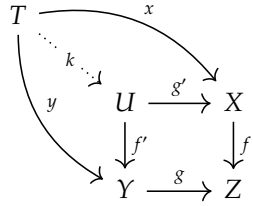

Figure~\ref{pullback}  illustrates the limit of a more complex diagram referred as a {\em {pullback}}, whose diagram is written abstractly as $\bullet \rightarrow \bullet \leftarrow \bullet$.  Note in Figure~\ref{pullback}, the functor maps the diagram $\bullet \rightarrow \bullet \leftarrow \bullet$ to actual objects in the category $Y \xrightarrow[]{g} Z \xleftarrow[]{f} X$. The universal property of the pullback square with the objects $U,X, Y$ and $Z$ implies that the composite mappings $g \circ f'$ must equal $g' \circ f$. In this example, the morphisms $f$ and $g$ represent a {\em {pullback}} pair, as they share a common co-domain $Z$. The pair of morphisms $f', g'$ emanating from $U$ define a {\em {cone}}, because the pullback square ``commutes'' appropriately. Thus, the pullback of the pair of morphisms $f, g$ with the common co-domain $Z$ is the pair of morphisms $f', g'$ with common domain $U$. Furthermore, to satisfy the universal property, given another pair of morphisms $x, y$ with common domain $T$, there must exist another morphism $k: T \rightarrow U$ that ``factorizes'' $x, y$ appropriately, so that the composite morphisms $f' \ k = y$ and $g' \ k = x$. Here, $T$ and $U$ are referred to as {\em cones}, where $U$ is the limit of the set of all cones ``above'' $Z$. If we reverse arrow directions appropriately, we get the corresponding notion of pushforward. 
\subsection{Category of Action-Value Functions is (co)Complete}

To illustrate diagrammatic reasoning in URL, we now prove that  the category ${\cal C_Q}$ is complete, that is, it contains all limits and colimits. Hence, all architecture diagrams are ``solvable", meaning that there is an action-value function that defines an object ``closest" to the diagram with respect to the morphisms coming into or out of the diagram.  Our proof hinges on two key properties: in the category ${\cal C}_{\bf Set}$ of sets, all diagrams are solvable, that is limits and colimits exist. Where possible, we reduce the problem of showing a property to that of the category ${\cal C}_{\bf Set}$. 
\begin{theorem}
\label{llmcomplete}
    The  category ${\cal C_Q}$ is {\em (co) complete}, meaning it contains all limits and colimits. 
\end{theorem}
{\bf Proof:}  Formally, this result requires showing that all diagrams, such as pullbacks $\bullet \rightarrow \bullet \leftarrow \bullet$, pushouts $\bullet \leftarrow \bullet \rightarrow \bullet$, (co)equalizer diagrams of the form $\bullet \rightarrow \bullet \twoarrows[0.4ex]{\rightarrow}{\rightarrow}  \bullet$ and $\bullet \twoarrows[0.4ex]{\rightarrow}{\rightarrow}  \bullet \rightarrow \bullet$, respectively, are solvable. For brevity, we will just illustrate the argument for pullback diagrams, and the other arguments are similar. Consider the cube shown in Figure~\ref{subobj-classifier}. Here, $f$, $g$, and $h$ are three value functions , each mapping some state into some real number. So, for example, the bottom face of the cube is a commutative diagram, meaning that $i \circ f = g \circ j$ holds. Similarly,  the arrows $p$, from  I" to  I', and arrow $q$ from O" to O' ensure the right face of the cube is a commutative diagram. The arrow from $P$ to $Q$ exists because looking at the front face of the cube, $Q$ is the pullback of $i$ and $q$, which must exist because we are in the category of sets ${\cal C}_{\bf Set}$, which has all pullbacks. Similarly, the back face of the cube is a pullback of $j$ and $p$, which is again a pullback in ${\cal C}_{\bf Set}$. Summarizing, $\langle u, v \rangle$ and $\langle m, n \rangle$ are the pullbacks of $\langle i, j \rangle$ and $\langle p, q \rangle$. The proof that ${\cal C_V}$ has all pushouts (limits) is similar. $\qed$

\section{Category of Action Value Functions form a Topos}
\label{topos-url} 

We now show that the category ${\cal C_Q}$ is not just (co)complete, but they have other properties that make them into a category called a topos \citep{maclane:sheaves,Johnstone:topostheory} that is a ``set-like" category with very special properties, which we will explore in the rest of the paper.  A topos generalizes all common operations on sets. The concept of subset is generalized to a {\em subobject classifier} in a topos.  To help build some intuition, consider how to define subsets without ``looking inside" a set. Essentially, a subset $S$ of some larger set $T$ can be viewed as a ``monic arrow", that is, an injective (or 1-1) function $f: S \hookrightarrow T$. 

\begin{definition} \citep{maclane:sheaves}
    An {\bf elementary topos} is a category ${\cal C}$ that has all (i) limits and colimits, (ii) has exponential objects, and (iii) a subobject classifier.
\end{definition} 
For example, the category of sets forms a topos. Limits exist because one can define Cartesian products of sets, and colimits correspond to forming set unions. Exponential objects correspond to the set of all functions between two sets. Finally, the subobject classifier is simply the subset function, which induces a boolean-valued characteristic function.  A few of the many alternative definitions are to combine (i) and (iii) by the condition that ${\cal C}$ is {\em Cartesian closed}; another reformulation is to replace condition (i) by the condition that ${\cal C}$ has a terminal object and pullbacks; and similarly, condition (ii) can be replaced by saying ${\cal C}$ has an initial object and coequalizers of all maps between coproducts. These are all different ways of characterizing a topos in terms of its universal properties. 
\begin{definition} 
    In a category ${\cal C}$ with finite limits, a {\bf subobject classifier} is a ${\cal C}$-object $\Omega$, and a ${\cal C}$-arrow ${\bf true}: {\bf 1} \rightarrow \Omega$, such that to every other monic arrow $S \hookrightarrow X$ in ${\cal C}$, there is a unique arrow $\phi$ that forms the following pullback square: 
%https://q.uiver.app/#q=WzAsNCxbMCwwLCJTIl0sWzMsMCwie1xcYmYgMX0iXSxbMCwyLCJYIl0sWzMsMiwie1xcYmYgMn0iXSxbMCwyLCJtIiwwLHsic3R5bGUiOnsidGFpbCI6eyJuYW1lIjoibW9ubyJ9fX1dLFswLDFdLFsxLDMsIntcXGJmIHRydWV9IiwxLHsic3R5bGUiOnsidGFpbCI6eyJuYW1lIjoibW9ubyJ9fX1dLFsyLDMsIlxccGhpX1MiLDEseyJzdHlsZSI6eyJib2R5Ijp7Im5hbWUiOiJkYXNoZWQifX19XV0=
\[\begin{tikzcd}
	S &&& {{\bf 1}} \\
	\\
	X &&& \Omega
	\arrow["m", tail, from=1-1, to=3-1]
	\arrow[from=1-1, to=1-4]
	\arrow["{{\bf true}}"{description}, tail, from=1-4, to=3-4]
	\arrow["{\phi}"{description}, dashed, from=3-1, to=3-4]
\end{tikzcd}\]  
\end{definition} 
This commutative diagram is a classic example of how to state a universal property: it enforces a condition that every monic arrow $m$ (i.e., every $1-1$ function) that maps a ``sub"-object $S$ to an object $X$ must be characterizable in terms of a ``pullback", a particular type of universal property that is a special type of a limit. In the special case of the category of sets, it is relatively easy to show that subobject classifiers are simply the characteristic (Boolean-valued) function $\phi$ that defines subsets. In general, as we show below, the subobject classifier $\Omega$ for causal models is not Boolean-valued, and requires using intuitionistic logic through a Heyting algebra.  This definition can be rephrased as saying that the subobject functor is representable. In other words, a subobject of an object $x$ in a category ${\cal C}$ is an equivalence class of monic arrows $m: S \hookrightarrow  x$. 

 \begin{theorem}
 \label{llmtopos}
     The category ${\cal C_Q}$  of value functions forms a topos. 
 \end{theorem}
 {\bf Proof:} The proof essentially involves checking each of the conditions in the above definition of a topos. We will focus on the construction of the subobject classifier and of exponential objects below, since we have already shown above in Theorem~\ref{llmcomplete} that the category ${\cal C_V}$ has all (co)limits. We prove each of these constructions in the next two sections. $\qed$

 \subsection{Subobject Classifiers for the category ${\cal C_Q}$ of action-value functions}

First, we need to define what a ``subobject" is in the category ${\cal C_Q}$ of action-value functions. Note that action-value functions   $f: S \times A \rightarrow \mathbb{R}$,  and $g: S' \times A' \rightarrow \mathbb{R}$ each map state-action pairs $s,a$, and $ s',a'$ into real numbers. Here, let us assume that the value function $f$ is a {\em sub-function} of the value function $g$. We can denote that by defining a commutative diagram as shown below. Note here that  $i$ and $j$ are monic arrows. Also, $O$ and $O'$ denote the set of possible values of the state-action pairs defined by sets $S \times A$ and $S' \times A'$, respectively. 
\begin{center}
\label{gdcsubobj}
% https://q.uiver.app/#q=WzAsNCxbMCwwLCJVIl0sWzIsMCwiVSciXSxbMCwyLCJWIl0sWzIsMiwiViciXSxbMCwxLCJpIiwwLHsic3R5bGUiOnsidGFpbCI6eyJuYW1lIjoibW9ubyJ9fX1dLFswLDIsImYiLDJdLFsxLDMsImciXSxbMiwzLCJqIiwyLHsic3R5bGUiOnsidGFpbCI6eyJuYW1lIjoibW9ubyJ9fX1dXQ==
\begin{tikzcd}
	S \times A && {S' \times A'} \\
	\\
	O && O'
	\arrow["i", tail, from=1-1, to=1-3]
	\arrow["f"', from=1-1, to=3-1]
	\arrow["g", from=1-3, to=3-3]
	\arrow["j"', tail, from=3-1, to=3-3]
\end{tikzcd}
\end{center} 
\begin{figure}
    \centering
    \includegraphics[width=0.4\linewidth]{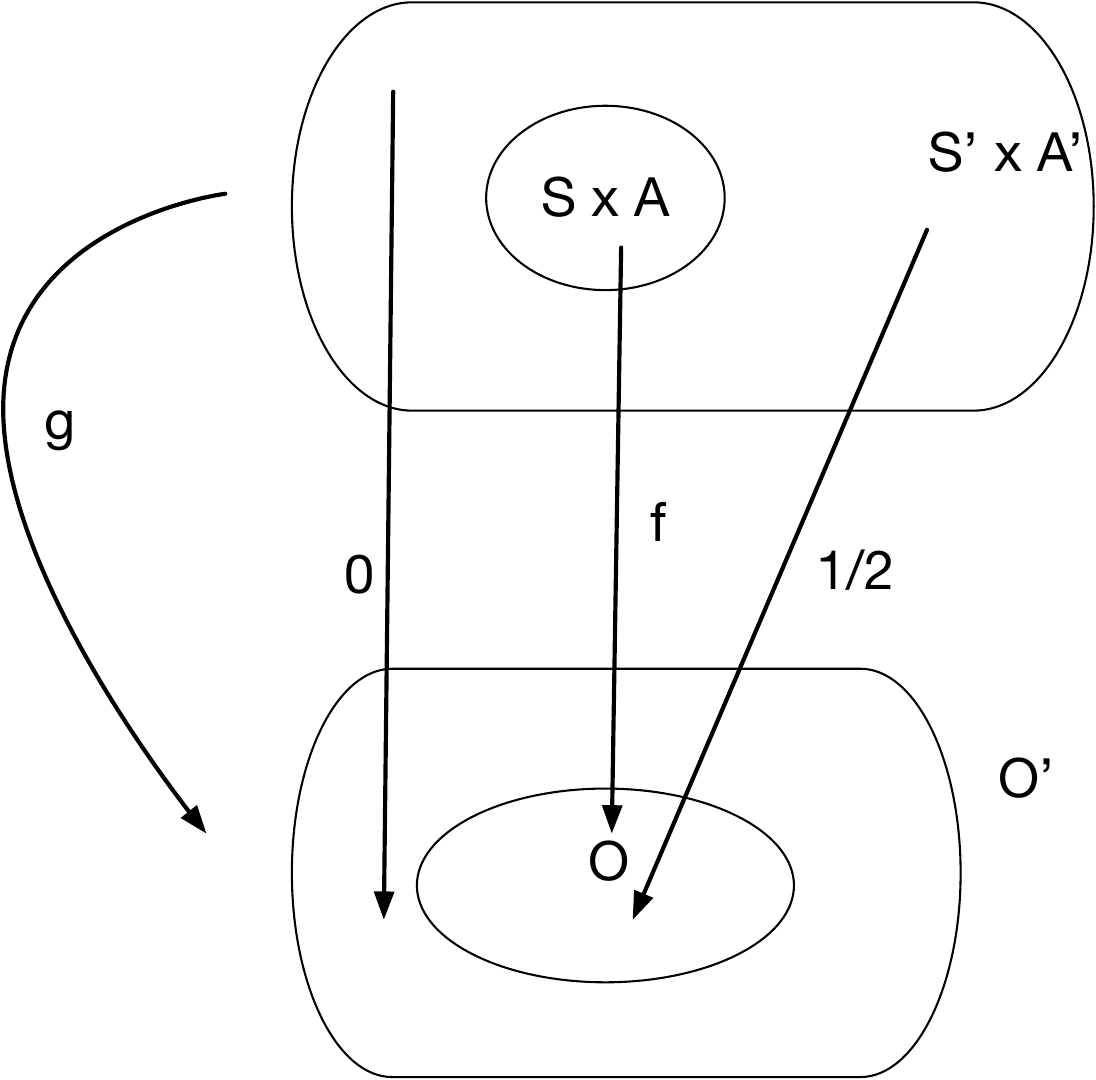}
    \caption{The subobject classifier $\Omega$ for the topos category ${\cal C_Q}$ of action-value functions. }
    \label{subobj}
\end{figure}
Let us examine Figure~\ref{subobj} to understand the design of subobject classifiers for the category ${\cal C_V}$. An element $x \in S' \times A'$, which is a particular state-action pair, can be classified in three ways by defining a (non-Boolean!) characteristic function $\psi$:
\begin{enumerate}
    \item $x \in S \times A$ -- here we set $\psi(x) = {\bf 1}$. 
    \item $x \notin S \times A$ but $g(x) \in O'$ -- here we set $\psi(x) = {\bf \frac{1}{2}}$.
    \item $x \notin S \times A$ and $g(x) \notin O$ -- we denote this by $\psi(x) = {\bf 0}$. 
\end{enumerate}

\begin{figure}
    \centering
    \includegraphics[width=0.4\linewidth]{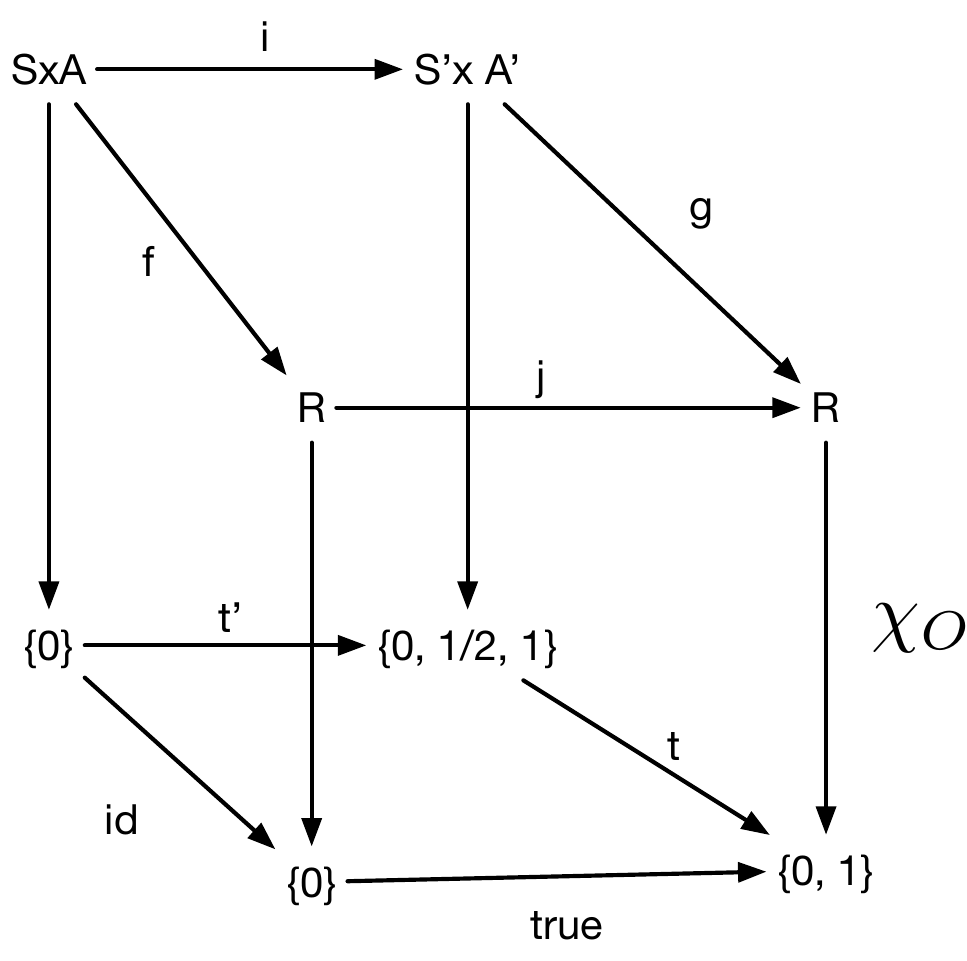}
    \caption{The subobject classifier $\Omega$ for the topos category ${\cal C_Q}$ of action-value functions is defined as the bottom face of the cube. }
    \label{subobj-classifier}
\end{figure}

The subobject classifier is illustrated as the bottom face of the cube shown in Figure~\ref{subobj-classifier}: 
\begin{itemize}
    \item ${\bf true}(0) = t'(0) = {\bf 1}$ 
    \item ${\bf t}: {\bf \{0, \frac{1}{2}, 1 \} \rightarrow \{0, 1 \}}$, where ${\bf t(0) = 0}, {\bf t(1) = t(\frac{1}{2}) = 1}$. 
    \item $\chi_O$ is the characteristic function of the output set $O$. 
    \item The base of the cube in Figure~\ref{subobj-classifier} displays the subobject classifier ${\bf T: 1 \rightarrow \Omega}$, where ${\bf T} = \langle {\bf t', true} \rangle$ that maps ${\bf 1 = id_{\{0\}}}$ to $\Omega = {\bf t}: \{0, \frac{1}{2}, 1 \} \rightarrow \{0, 1\}$.
\end{itemize}

This proves that the subobject classifier exists for the category ${\cal C_V}$ of value functions,  and it is not Boolean, but has three values of ``truth" , corresponding to the three types of classifications of monic arrows (in regular set theory, subobject classifiers are Boolean: either an element of a parent set is in a subset, or it is not).  

\subsection{Exponential Objects in the Category ${\cal C_Q}$}

To complete the proof of Theorem~\ref{llmtopos}, we need to prove also that the category ${\cal C_Q}$ has ``exponential objects". Given action value functions $f: S \rightarrow O$, and $g: S \rightarrow O'$, where $S, S'$ refer to state-action sets, and $O, O'$ refer to subsets of Euclidean space (real numbers), we define the {\em exponentiated value function} $g^f: E \rightarrow F$, where $F = O'^O$ is the regular exponential object in the category ${\cal C}_{\bf Set}$ (i.e., all functions from the set $O$ to $O'$),  and $E$ is the collection of all arrows from the value function $f$ to value function $g$ in the category ${\cal C_V}$,  which can be written more precisely as
\[ E = \{ \langle h, k \rangle | h, k \ \ \mbox{are arrows in the diagram below} \}\]
\begin{center}
%\label{expo-llm}
%https://q.uiver.app/#q=WzAsNCxbMCwwLCJVIl0sWzAsMiwiViJdLFsyLDAsIlUnIl0sWzIsMiwiViciXSxbMCwxLCJmIl0sWzAsMiwiaCIsMl0sWzIsMywiZiciLDJdLFsxLDMsImciXV0=
\begin{tikzcd}
	S && {S'} \\
	\\
	O && {O'}
	\arrow["h"', from=1-1, to=1-3]
	\arrow["f", from=1-1, to=3-1]
	\arrow["{g}"', from=1-3, to=3-3]
	\arrow["k", from=3-1, to=3-3]
\end{tikzcd}
 \end{center}
 and $g^f(\langle h, k \rangle) = k$. First, we define the ``product" object of $g^f$ and $f$ in the category ${\cal C_V}$ as the product map $g^f \times f: E \times I \rightarrow F \times O$. To show that we have defined a genuine exponential object, we need to demonstrate an ``evaluation" map $g^f \times f \rightarrow g$. The evaluation arrow from $g^f \times f$ to $g$ is the pair $\langle u, v \rangle$ defined by the following commutative diagram:
\begin{center}
 \label{expo-llm}
%https://q.uiver.app/#q=WzAsNCxbMCwwLCJVIl0sWzAsMiwiViJdLFsyLDAsIlUnIl0sWzIsMiwiViciXSxbMCwxLCJmIl0sWzAsMiwiaCIsMl0sWzIsMywiZiciLDJdLFsxLDMsImciXV0=
\begin{tikzcd}
	E \times I && {I'} \\
	\\
	F \times O && {O"}
	\arrow["u"', from=1-1, to=1-3]
	\arrow["g^f \times f", from=1-1, to=3-1]
	\arrow["{g}"', from=1-3, to=3-3]
	\arrow["k", from=3-1, to=3-3]
\end{tikzcd}
 \end{center}
Here, $v$ is the usual evaluation arrow in the category ${\cal C}_{\bf Set}$ of sets, and $u$ maps $\langle \langle h, k \rangle, x \rangle$ to output $h(x)$. 

\subsection{More General Topos Categories in URL}

For simplicity, we worked through the details of the proof that the category ${\cal C_Q}$ of action-value functions forms a topos, but a similar proof can be constructed for a variety of other categories for URL. For example, if we consider the MDP and PSR homomorphisms defined in Section~\ref{urlcat}, we can define a topos category ${\cal C}_{MDP}$ and  a topos category ${\cal C}_{PSR}$ as well, where we need to follow the same procedure described above for constructing subobject classifiers and exponential objects. MDP subobjects would correspond to parts of another MDP, which have been the topic of extensive work in hierarchical RL \citep{hrl}. 

\section{Summary and Future Work}

In this paper, we described Universal Reinforcement Learning (URL),  a categorial generalization of RL, which builds on abstractions from the study of coinduction on non-well-founded sets and universal coalgebras, topos theory, and categorial models of asynchronous parallel distributed computation.  In the first half of the paper, we  illustrated how categories and functors can be defined in the basic RL framework.  We modeled the  space of algorithms for MDPs or PSRs as a functor category, where the co-domain category forms a topos, which admits all (co)limits, possesses a subobject classifier, and has exponential objects. In the second half of the paper, we described  a broad family of universal coalgebras, extending the dynamic system models studied previously in RL. We introduced the core problem of finding the final coalgebra, and reviewed Lambek's theorem.   Some suggested directions for future work are described below. 

\begin{enumerate}
    \item {\em Stochastic approximation as Kan extensions}: One popular approach for analyzing the convergence of RL algorithms is through the use of {\em martingales}. Recent advances in categorical probability theory have shown that the Martingale Convergence Theorem can be viewed as a type of Kan extension \citep{rubenvanbelle:phd}. The application of Kan extensions to RL is an area that is worth investigating in more detail. 

    \item {\em Primal-dual space RL:} We restricted our analysis of URL to the simplest setting where no function approximation was involved. We proposed a broad framework of proximal RL \citep{mahadevan2014proximal} previously, and it would be worth exploring a proximal URL framework as well. 

    \item {\em Deep URL:} We briefly showed in Section~\ref{deeprl} how to model deep RL as a functor, building on the work of \citep{DBLP:conf/lics/FongST19}. Much more needs to be done here in exploring the compositional structure of deep URL. 
\end{enumerate}

\newpage 
% References
%\bibliography{allcitations}

\end{document}